\documentclass[10pt,twocolumn,letterpaper]{article}
\usepackage[pagenumbers]{cvpr} 

\usepackage[table,xcdraw]{xcolor}
\usepackage{amsmath}
\usepackage{amssymb}
\usepackage{float}
\usepackage{overpic}
\usepackage[export]{adjustbox} 
\newcommand\laura[1]{\NOTE{red}{Laura}{#1}}
\newcommand\NOTE[3]{\textcolor{#1}{[#2: #3]}}
\definecolor{jorange}{rgb}{8,.5,.0}
\newcommand\linus[1]{\NOTE{jorange}{Linus}{#1}}
\definecolor{bluegreen}{rgb}{.05,.6,.7}
\newcommand\marc[1]{\NOTE{magenta}{Marc}{#1}}
\definecolor{grey}{rgb}{.6,.6,.6}
\newcommand\BE[1]{\NOTE{magenta}{Bernhard}{#1}}
\newcommand\was[1]{\NOTE{grey}{Was}{#1}}

 \renewcommand\marc[1]{}
 \renewcommand\linus[1]{}
 \renewcommand\laura[1]{}
 \renewcommand\was[1]{}
 \renewcommand\BE[1]{}
 
\newcommand{\lossgeo}{\mathcal{L}_{\textit{geo}}}
\newcommand{\lossmse}{\mathcal{L}_\textit{photo}}
\newcommand{\regpoisson}{\mathcal{R}_{\textit{p}}}
\newcommand{\regedge}{\mathcal{R}_{\textit{e}}}
\newcommand{\regsmooth}{{\mathcal{R}_{\textit{s}}}}
\newcommand{\lossgeomconsistency }{{\mathcal{L}_{\textit{gc}}}}

\newcommand{\I}{\mathbf{I}}
\newcommand{\D}{\mathbf{D}}
\newcommand{\M}{\mathbf{M}}
\newcommand{\X}{\mathbf{X}}
\newcommand{\E}{\mathbf{E}}

\newcommand\arxivorelse[2]{#2} 
\renewcommand\arxivorelse[2]{#1} 
\definecolor{cvprblue}{rgb}{0.21,0.49,0.74}
\usepackage[pagebackref,breaklinks,colorlinks,citecolor=cvprblue]{hyperref}

\title{Refinement of Monocular Depth Maps via  Multi-View Differentiable Rendering}

\author{
\begin{tabular}{ccccc}
Laura Fink$^{1,2}$
&
Linus Franke$^{1}$
& 
Bernhard Egger$^{1}$%
&
Joachim Keinert$^{2}$
&
Marc Stamminger$^{1}$
\end{tabular}
\\
\and
$^{1}\label{tt}$Friedrich-Alexander-Universit\"at\\
Erlangen-N\"urnberg\\
{\tt\small \{first.last\}@fau.de}
\and
$^{2}$Fraunhofer IIS\\
Erlangen\\
{\tt\small \{first.last\}@iis.fraunhofer.de}
}

\begin{document}

\twocolumn[{ 
\renewcommand\twocolumn[1][]{#1} 
\maketitle 
\centering
    \vspace{-0.3cm}
    \includegraphics[width=.9\textwidth]{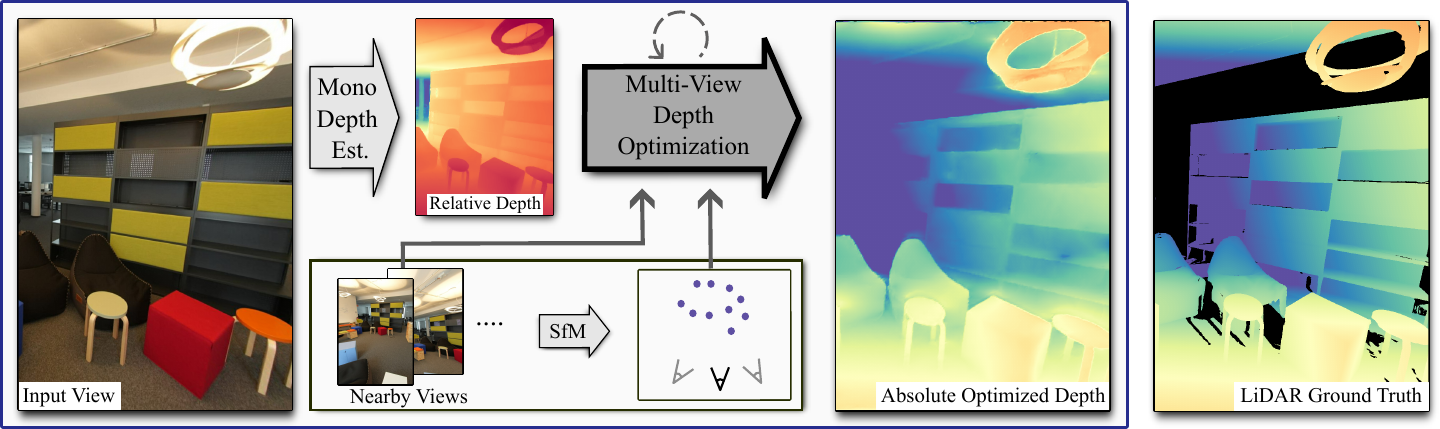}
    \label{fig:teaser} 
    \captionof{figure}{Our novel algorithm refines a depth map from a monocular estimator with multi-view differentiable rendering based on a meshed representation of the target depth map. This results in dense and accurate depth maps, especially in challenging indoor scenarios. 
    Note that the ground truth here is obtained from a meshed LiDAR scan, as such uncertain areas are left blank.}  
    \vspace{0.35cm} 
}] 

Accurate depth estimation is at the core of many applications in computer graphics, vision, and robotics. 
Current state-of-the-art monocular depth estimators, trained on extensive datasets, generalize well but lack 3D consistency needed for many applications. 
In this paper, we combine the strength of those generalizing monocular depth estimation techniques with multi-view data by framing this as an analysis-by-synthesis optimization problem to lift and refine such relative depth maps to accurate error-free depth maps. 
After an initial global scale estimation through structure-from-motion point clouds, we further refine the depth map through optimization enforcing multi-view consistency via photometric and geometric losses with differentiable rendering of the meshed depth map.
In a two-stage optimization, scaling is further refined first, and afterwards artifacts and errors in the depth map are corrected via nearby-view photometric supervision.
Our evaluation shows that our method is able to generate detailed, high-quality, view consistent, accurate depth maps, also in challenging indoor scenarios, and outperforms state-of-the-art multi-view depth reconstruction approaches on such datasets.
\\
Project page and source code can be found at \\
{\small \url{https://lorafib.github.io/ref_depth/}}.


\section{Introduction}


Depth estimation is crucial in computer vision, graphics, and robotics~\cite{dong2022towards}. 
They support online use cases like autonomous driving~\cite{wang2019pseudo}, augmented reality~\cite{kalia2019real}, or robot navigation~\cite{dong2022towards, kalia2019real},
but also many offline scenarios like 
3D reconstruction~\cite{laga2020survey},
supervision for 3D foundation models like DUSt3r or MASt3R~\cite{leroy2024grounding}, post-processing for computational photography and
video editing (defocus blur, background removal, etc.), content creation for light field displays,   or virtual scene exploration~\cite{hedman2018deep, fink2023livenvs}.
In this paper, we focus on building a \textit{stable} optimization pipeline that favors highest fidelity of \textit{individual} depth maps for such non-realtime use cases that especially profit from dense and accurate depth information.

\begin{figure}
\centering
\includegraphics[width=.98\linewidth]{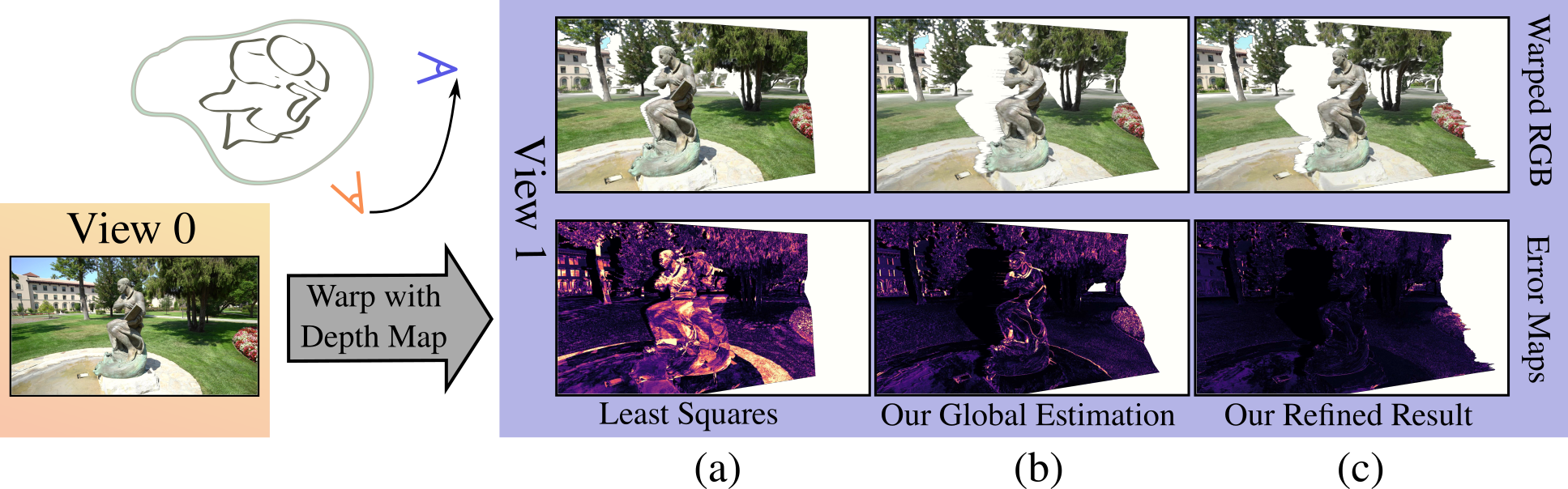}
\caption{Warping with depth maps reveals inconsistencies and artifacts. Absolute depth maps~\cite{yang2024depth} estimated with least squares display missalignments upon warping.  In contrast, our approach, which combines estimation and optimization, effectively preserves and enforces multi-view consistency.}
\label{fig:motivation}
\end{figure}

For the mentioned scenarios, most often depth maps are used that have been generated by patch-matching methods like COLMAP \cite{schonberger2016structure} or  MVSNets~\cite{yao2018mvsnet}, which often suffer from incomplete reconstructions.
Alternatively, recent pre-trained monocular depth estimation methods~\cite{ge2024geobench, ke2024repurposing, yang2024depth2, yang2024depth} provide good and complete results, although mostly in the form of \textit{relative} depth maps in a normalized scale.
Transferring these depth maps to absolute values is possible via least squares fitting to LiDAR data~\cite{ke2024repurposing,ranftl2020towards}, or with sparse, absolute multi-view point cloud data~\cite{yao2018mvsnet} as obtained by Structure-from-Motion (SfM)~\cite{schonberger2016structure}.
Yet, this approach also poses problems due to unreliable min/max points and uneven SfM point cloud distributions (see Fig.~\ref{fig:motivation}~(a)).

\begin{figure*}[t]
\centering

 \hfill
 \begin{overpic}[width=.23\linewidth, valign=t]{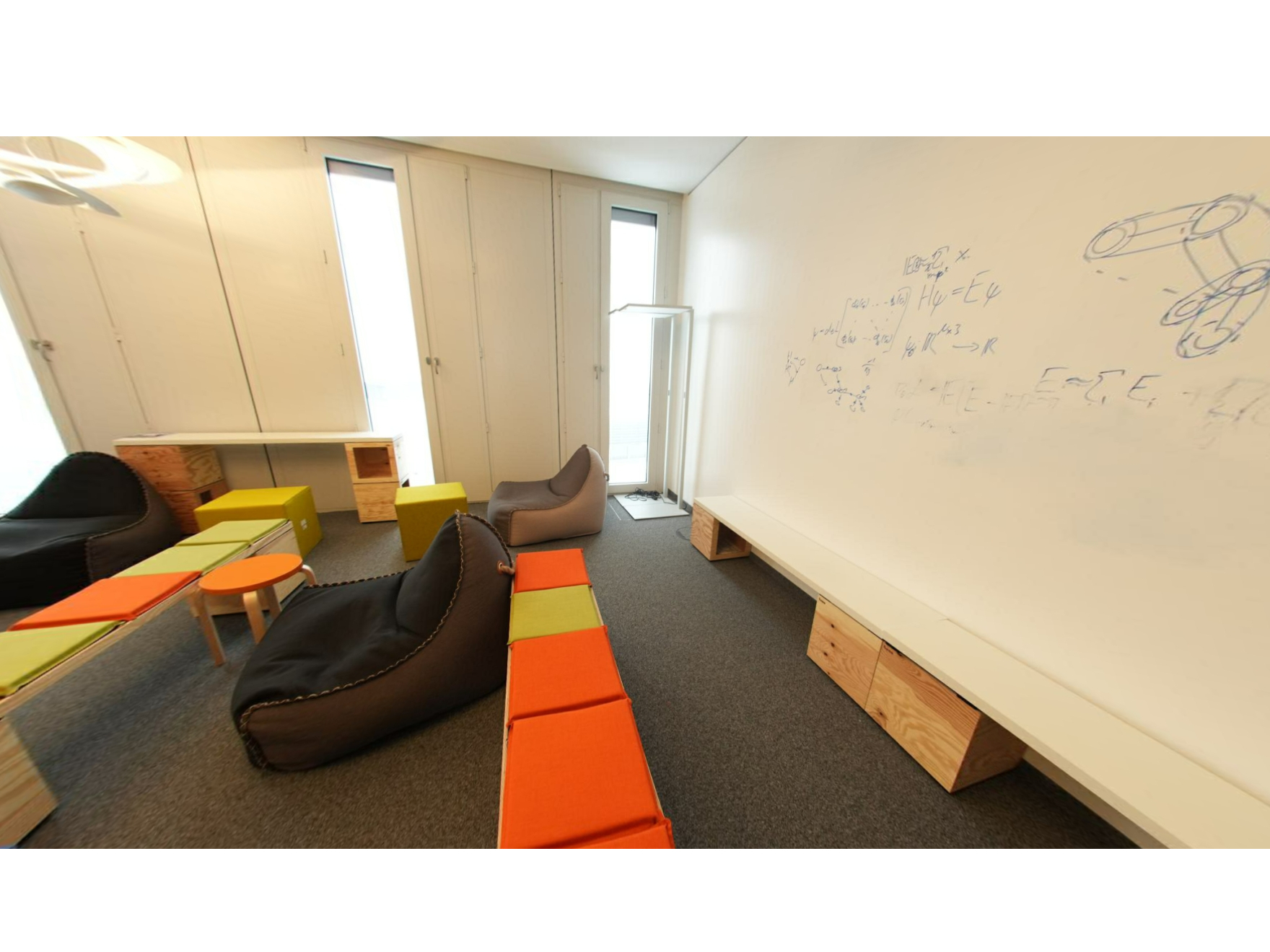}
 \put(50,3){\makebox(0,0){\tiny Input Image}}
 \end{overpic}
 \hfill
\includegraphics[width=0.69\linewidth]{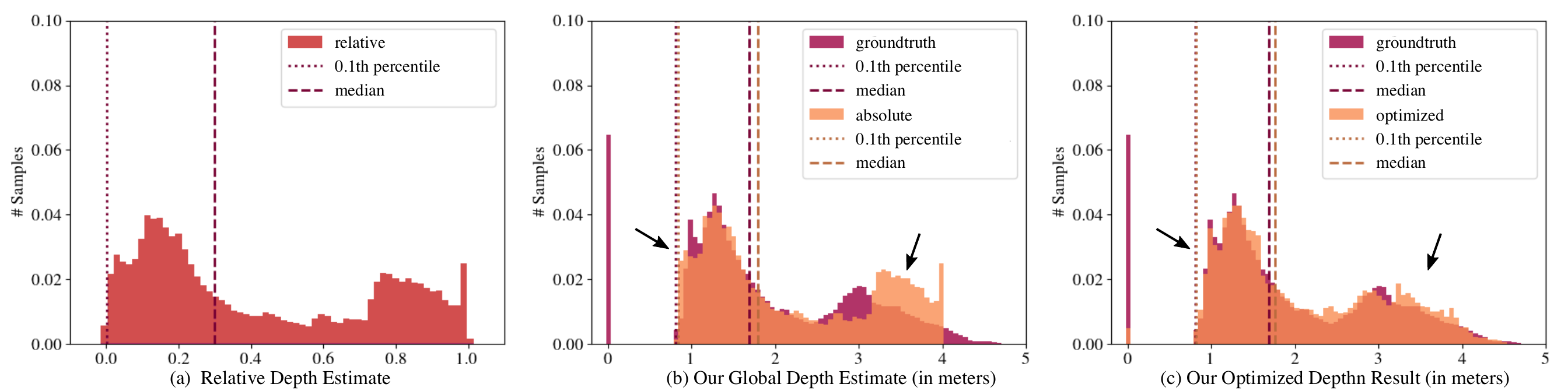}
 \hfill
\caption{Histograms of different states of depth maps during our optimization. The initial estimate (a) is in relative space (red) and taken as input to be refined (orange). Our absolute estimation (b) brings this to absolute space by aligning the medians and 0.1th percentiles. Then, we  optimize (c) to closely capture the depth distribution of the ground truth (red). Missing values, like windows, are counted as 0\,m.
}
\label{fig:histograms}
\end{figure*}

The approach proposed in this paper follows this idea to rescale estimated depth maps to absolute values using SfM data.
It is based on two observations:
First, similar to the Multi-view Stereo (MVS) community~\cite{yao2018mvsnet}, where bounding volumes are guided by SfM, we find that the scene extent can be determined by aligning the medians of the depth map and sparse point cloud leading to a robust initial alignment (see Fig.~\ref{fig:motivation}~(b)). 
Second, we find that explicitly utilizing SfM data and multi-view images and combining it with a differentiable renderer can significantly enhance \textit{scale refinement} and enable \textit{error correction} of depth map inaccuracies (see Fig.~\ref{fig:motivation}~(c)). 
These errors, typically manifesting as noise, oversmoothing or distance errors between objects, change the relative alignments in a depth map, as seen in the histogram in Fig.~\ref{fig:histograms} (b).
By \textit{warping} the input image to adjacent views using the depth map and performing a comparison, we observe that depth errors become highly apparent (Fig.~\ref{fig:motivation}). 
Consequently, we propose using a differentiable nearby-view rendering framework, which enables an optimization strategy tailored to address this challenge.

Our pipeline works as follows: First, we use a extensively trained monocular depth estimator~\cite{ke2024repurposing, yang2024depth2} to predict individual image depths, generating smooth but non-metric depth maps.
As mentioned before, we estimate our global scale and convert the depth map into a triangle mesh for rendering and optimization.

Then we optimize the resulting depth mesh towards a finely detailed multi-view consistent absolute depth map using a differentiable renderer~\cite{Laine2020diffrast}.
Optimization takes place in two stages.
First, we learn a neural field that remaps the depth map mesh coarsely to the correct values, fitting it directly against the SfM point cloud.
This efficiently mitigates distance errors.
Second, we apply local depth map refinement in an analysis-by-synthesis paradigm, supervising warped-colored renderings with the captured input images.
This removes oversmoothing and restores fine details in the depth map.

As we show in our evaluation on synthetic and real-world data, our method provides highly detailed and accurate depth maps, surpassing competing methods.

Our contributions are as follows.
\begin{itemize}
    \item A novel analysis-by-synthesis technique to refine monocular depth maps for accurate 3D information via view consistency optimization.
    \item A two-step refinement scheme that first performs a coarse alignment via shallow neural fields and second follows a local refinement strategy to optimize depth values on fine granularity for highly detailed depth maps.
    \item An edge aware and a Poisson blending~\cite{perez2003poisson} inspired regularizer to exploit the strong initial estimates from monocular estimators. Yielding robust results even in challenging scenarios.
    \item An extensive evaluation of the proposed method, showcasing its efficiency especially in difficult feature-scarce indoor scenes.
\end{itemize}

\section{Related Work}

\newcommand{\relworksubsection}[1]{\paragraph*{#1}}

\begin{figure*}[t]
\centering
\begin{overpic}[width=1\linewidth,keepaspectratio]{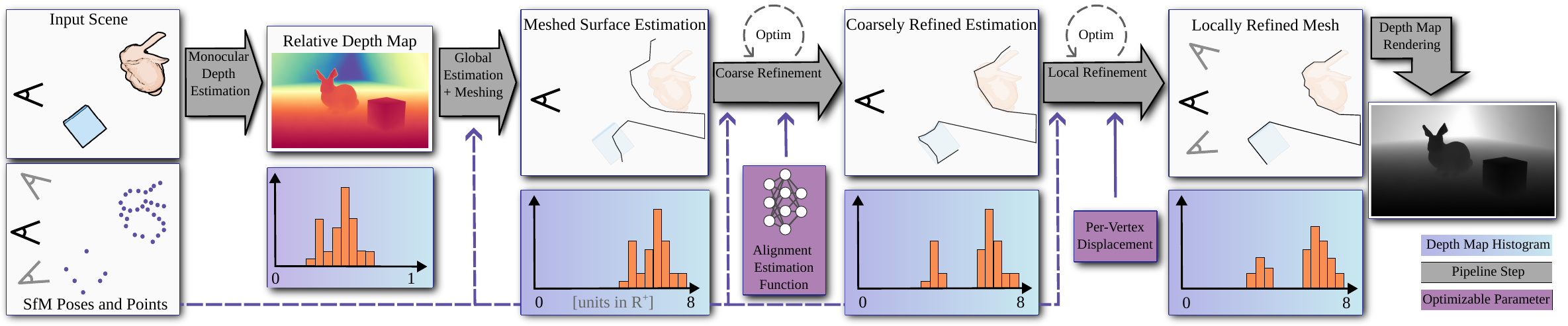}
 \put(14,13.85){\makebox(0,0){\fontsize{4.0}{4.0}\selectfont(Sec.~\ref{sec:mono_estimate})}}
 \put(30.2,13.85){\makebox(0,0){\fontsize{4.0}{4.0}\selectfont(Sec.~\ref{sec:meshingsec})}}
 \put(49.2,14.8){\makebox(0,0){\fontsize{4.6}{4.6}\selectfont(Sec.~\ref{sec:coarse_refinement})}}
 \put(70,14.8){\makebox(0,0){\fontsize{4.6}{4.6}\selectfont(Sec.~\ref{sec:local_refinement})}}
 \put(26,8.8){\makebox(0,0){\scriptsize $\mathbf{D}_0^r$}}
 \put(43.8,7.4){\makebox(0,0){\scriptsize $\mathbf{D}_0^g$}}
 \put(64.8,7.4){\makebox(0,0){\scriptsize $\mathbf{D}_0^c$}}
 \put(85.5,7.4){\makebox(0,0){\scriptsize $\mathbf{D}_0$}}
\end{overpic}
    \vspace{-0.5cm}
\caption{Overview of our method. We employ monocular depth estimation for a relative, but topologically complete depth map. Results from Structure-from-Motion are used to scale the depth map to absolute space. Following, we convert the depth map to a surface mesh for refinement via differentiable rendering. The refinement is done in two consecutive steps: first, we learn a mapping function that smoothly aligns the depth map to the sparse point cloud and second, we refine per-vertex positions, yielding accurate depth maps.}
\label{fig:pipeline}
\end{figure*}

\relworksubsection{Image-based Matching \& Triangulation.}

Multi-view-stereo (MVS)~\cite{hartley2003multiple, schoenberger2016mvs} methods integrate data from multiple images by matching and triangulating patches based on horizontal disparity~\cite{bleyer2011patchmatch}. 
This resembles stereo pair patch matching along the epipolar line to increase match probability through similarity computations.
Scarce textures pose challenges, causing sparse results in such areas. Coarse-to-fine strategies~\cite{xu2019multi, wang2023adaptive} or plane prior methods have been implemented to address these issues~\cite{xu2022multi, yuan2024tsar, wang2023adaptive}.

The plane sweep algorithm~\cite{collins1996space} is a method for estimating MVS depth by dividing the viewing area into frontoparallel planes at varying depths and warping them into other views. 
It projects colors from all views into the camera's frustum, choosing the depth hypothesis with optimal photometric consistency. 
The quality of depth estimates heavily depends on the number of planes, or essentially the resolution of this volumetric representation.

Yao et al.~\cite{yao2018mvsnet} combined the plane sweep approach with neural networks. 
Several follow-up works refined the initial architecture and training~\cite{ding2022transmvsnet, liao2022wt, liu2023epipolar, cao2024mvsformer++, Cao2022MVSFormerMS, gu2020cascade,cheng2020deep,yang2020cost, yao2019recurrent,wei2021aa,yan2020dense, zhu2021deep}.
To date, these networks dominate popular benchmarks such as the Tanks and Temples~\cite{knapitschTanksTemplesBenchmarking2017a} point cloud reconstruction benchmark.

\relworksubsection{Novel View Synthesis \& Inverse Rendering.}
In the realm of novel view synthesis~(NVS), depth maps serve as a common way to represent an underlying 3D scene~\cite{fink2023livenvs, hedman2018deep}.
Thus, some methods started to produce depth maps within their pipeline as a byproduct to enable high quality NVS~\cite{wang2021ibrnet,zhang2023structural, wizadwongsa2021nex}.
However, these depth maps often have very limited depth precision~\cite{wizadwongsa2021nex} or resolution~\cite{wang2021ibrnet}.

Others use advancements from the novel view synthesis  and inverse rendering community to directly optimize for depth maps.
Kim et al.~\cite{kimmc2} integrate image-based rendering methods~\cite{li2023dynibar} to infer view-consistent depth maps. 
Chugunov et al.~\cite{chugunov2023shakes} transfer multi resolution hash grids~\cite{muller2022instant} to optimize near-field depth maps from RAW camera bursts.
Mirdehghan et al.~\cite{mirdehghanTurboSLDenseAccurate} and Shandilya et al.~\cite{shandilya2023neural} cast structured-light scanning as an inverse rendering problem.

\relworksubsection{Refinement.}
To mitigate problems in depth maps, such as sparseness and noise, various methods aim to improve low-quality maps by using additional information or integrating data from other sensors, such as aligned photographs. 
Shape-from-shading techniques estimate the curvature of light-surface interactions, using registered color data to refine the quality of the depth map~\cite{tao2015depth, haefner2018fight, kadambi2017depth, or2016real, wu2014real, or2015rgbd}. 
Some also optimize the photometric consistency between views with NeRF-based schemes~\cite{ito2024neural}.

Other methods~\cite{zhang2020listereo, ma2019self, gruber2019gated2depth, park2024flexible} use neural networks to complete depth maps from sparse data such as LiDAR of SfM point clouds. 
Similarly, low-resolution depth maps are upsampled using shape-from-shading~\cite{haefner2018fight} or temporal information~\cite{yuan2018temporal, sun2023consistent}.
Other approaches tackle noise~\cite{sterzentsenko2019self, yan2018ddrnet} in depth maps.

\relworksubsection{Monocular Depth Estimation.}
Monocular depth estimators~\cite{ke2024repurposing, yang2024depth, spencer2024third, ornek20222d} take shading-based refinement and image-based depth densification further by inferring 3D structures from a single image. 
Many methods already deliver depth maps with pixel-perfect silhouettes and close approximations of curvatures.
However, these depth estimators are geometrically inaccurate and ambiguous due to their lack of geometric triangulation and struggle with ambiguous scenes~\cite{ge2024geobench, ornek20222d}.
Using their learned priors, these estimators can be integrated into multi-view systems~\cite{kimmc2} or for the regularization of related tasks~\cite{li2024dngaussian}.

\section{Method}

We aim to build a pipeline which provides a highly detailed and multi-view consistent depth map for a \textit{set} of posed input images, see Fig.~\ref{fig:pipeline} for a visualization.

The pipeline begins with a relative depth map from monocular depth estimation. 
We then scale this non-metrical estimation to absolute values (refer to Secs.~\ref{sec:input} and \ref{sec:mono_estimate}). 
This adjustment uses the SfM point cloud derived from nearby views. 
However, the map still exhibits local view inconsistencies (see Fig.~\ref{fig:motivation}).
We transform the depth map into a fine-grained triangle depth mesh, which is then optimized using inverse rendering, incorporating other images as RGB supervision.

This optimization is performed in two steps:
Firstly, we do a \textit{coarse refinement} with only a few degrees of freedom via a very shallow coordinate MLP, which optimized the scaling and offset for the depth map (see Sec.~\ref{sec:coarse_refinement}).
In the second step, we \textit{locally} refine (see Sec.~\ref{sec:local_refinement}) the depth mesh on a per-vertex basis using differentiable rendering (see Sec.~\ref{sec:diffrast}) aiming for photo-consistency with the other images.
To enable robust optimization even in varying lighting conditions, we further integrate a differential tonemapping module (see the \arxivorelse{Appendix~\ref{suppl:tonemapping}}{supplemental material}).

\subsection{Input}
\label{sec:input}

$\mathbf{I}_i\in \mathbb{R}^{H\times W \times 3}, i \in\{0,\ldots,n\}$, denotes one of $n$ input images, with $\mathit{W}$ and $\mathit{H}$ being the images' width and height.
We first use SfM, specifically COLMAP~\cite{schonberger2016structure}, to estimate intrinsics $\mathbf{C}_i$ and pose $\{(\mathbf{R},\mathbf{t})_i\}$ of each image, as well as a sparse point cloud $\X=\{\mathbf{x}_j\}$ with all $\mathbf{x}_j$ being visible in $\I_0$.
$\mathbf{C}$ is defined with the common convention of focal lengths $(f_x,f_y)$ and principal point $(c_x,c_y)$ and images are undistorted.

\subsection{Monocular Depth and Global Estimation}\label{sec:mono_estimate}

We estimate a depth map $\D_0^r \in \mathbb{R}^{H\times W \times 1} $ for $\I_0$ using a strongly pretrained monocular depth estimator.

To approximately map the relative depth map $\D_0^r$ to an absolute scale, we use the sparse point cloud $\X$:
we globally offset and scale $\D_0^r$ such that its median distance and the value of its 0.1th percentile are aligned with the corresponding values of the sparse point cloud $\X$, resulting in depth image $\D_0^g$.

Note that we also considered to refine multiple depth maps in parallel (see \arxivorelse{Appendix~\ref{suppl:ablation_studies}}{the supplemental} for more details).
However, we opted to not do so as the resulting differences are negligible while the VRAM drastically increases. 

\subsection{Meshing}
\label{sec:meshingsec}

Next, we convert $\D_0^g$ to a 3D depth mesh $\M$.
We opted for a meshed representation as it can be trivially initialized from any dense depth estimation.
Furthermore, meshes allow for a non-uniform representation of the underlying geometry.
In particular, we can use reduction techniques that further regularize our results and reduce resource consumption.

\paragraph*{Triangle mesh.}
We unproject depth samples at image position $(u,v)$ with $ u \in \left[0,W-1\right]$ and $ v \in \left[0,H-1\right]$ to obtain a vertex $\textbf{p} = (x,y,z)^T$ with
\begin{equation}
\small 
    x = (u-c_x)\frac{z}{f_x},  \quad y = (v-c_y)\frac{z}{f_y}.
\label{eq:pointofmesh}
\end{equation}
and connect the vertices via triangle-based quads.


We usually use a mesh resolution of $W/d \times H/d$ triangle quads, with $d=4$ to avoid sub-pixel sized triangles, which do not receive gradients during differentiable rasterization and may lead to outliers (see Sec.~\ref{sec:diffrast}).
Although this seems as constraining resolution, note that the default lower resolution of estimators~\cite{ke2024repurposing} equates to $d$ between 2.3 and 2.6 and furthermore that $(u,v)$ can shift within a quad, facilitating piecewise linear silhouette and edge approximation.

The mesh colors are taken from $\I_0$ using the interpolated image coordinates $(u,v)$, separating geometric complexity from image resolution and allowing for vertex reduction.

\begin{figure}[t]
\centering
\includegraphics[width=.98\linewidth]{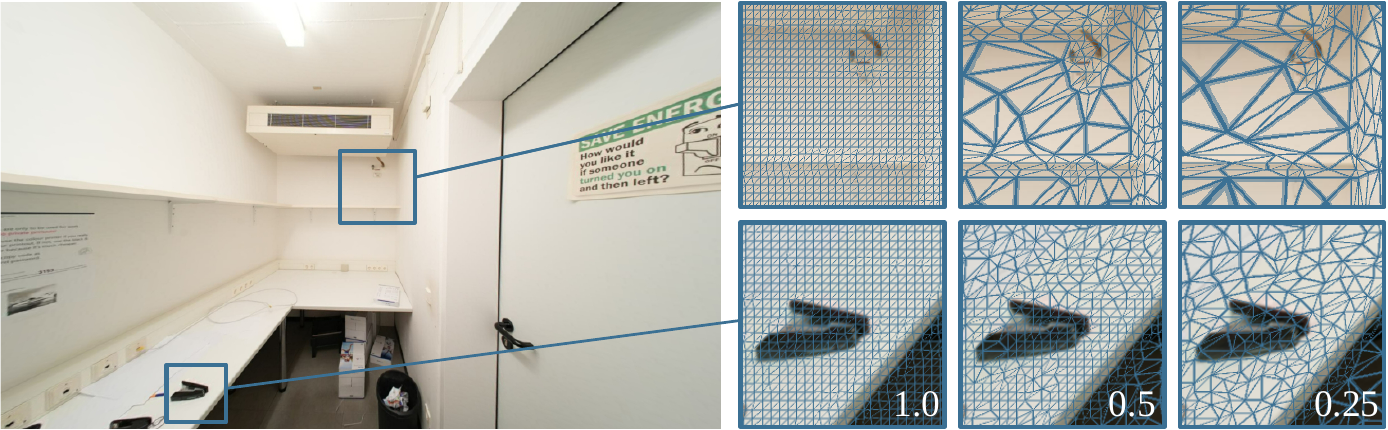}
\caption{Visualization of our depth mesh $\M$, where triangle size varies with scene complexity. Numbers in lower left corner indicate decimation ratio $r$. }
\label{fig:decimation}
\end{figure}

\paragraph*{Decimation.}
Scene geometry frequently features flat surfaces. 
Standard mesh simplification techniques identify these areas and replace smaller triangles with larger ones, simplifying the mesh. 
From the initial depth estimation, planar surfaces can be anticipated, allowing us to automatically allocate optimizable parameters strategically. 
The decimation process naturally regularizes refinement, enhancing robustness even in complex scenes with numerous sparse-feature walls, as demonstrated in Fig.~\ref{fig:decimation}. 
This can be likened to the plane prior support used in patch match-based approaches~\cite{xu2022multi}.

We employ quadric mesh decimation~\cite{garland1997surface} on unprojected mesh vertices before local refinement~(refer to Sec.~\ref{sec:local_refinement}).
Decimating vertices in non-linear space~(\arxivorelse{as per Appendix~\ref{suppl:nvdiffrast_support}}{see the supplement}) favors a vertex density near the camera.
Quadric mesh decimation efficiently maintains position accuracy, avoiding erroneous non-linear interpolation.

\subsection{Differentiable Rendering}
\label{sec:diffrast}
We use a differentiable render function $\Phi$ defined as
\begin{equation}
    (\I^*_i,\D^*_i) = \Phi(\textbf{C}_i, \textbf{R}_i, \textbf{t}_i, \textbf{C}_0, \textbf{R}_0, \textbf{t}_0, \M).
    \label{eq:render}
\end{equation}
$\Phi$ projects the current mesh $\M$ with camera parameters $\textbf{R}_0,\textbf{t}_0,\textbf{C}_0$ to a neighboring view $i$ with $\textbf{R}_i,\textbf{t}_i,\textbf{C}_i$, creating a pair of synthesized RGB images and depth maps.
Specifically, all vertices $\textbf{p} \in \M$ (see Eq.~\ref{eq:pointofmesh}) are projected with 
\begin{equation}
\label{eq:rendering_projection}
    \textbf{p}^*_i = \textbf{C}_i [\textbf{R}_i|\textbf{t}_i]  [\textbf{R}_0|\textbf{t}_0]^{-1} \textbf{p}
\end{equation}
and then rasterized. 
During optimization, the state of $\M$ continuously changes as described in Sec.~\ref{sec:coarse_refinement} and \ref{sec:local_refinement}.
In the following, we denote the current state of the depth map as $\D_0^*$.
Returning from the  meshed representation $\M$ to $\D_0^*$ is straight forward by evaluating $\Phi$ with $i=0$, thus $\textbf{C}_0,\textbf{R}_0,\textbf{t}_0$, which also yields $\mathbf{I}_0^*$, both of which are rasterized versions of $\M$.

\subsection{Coarse Refinement}\label{sec:coarse_refinement}

In the first optimization stage, the goal is to transfer the initial depth estimation~$\D_0^g$ to more accurate values.
Note that in this stage, we do not aim for highest precision, but mainly to coarsely align our depth buffer to the sparse SfM point cloud having only a few degrees of freedom.

To this end, we determine a neural field that yields an offset~$o$ and a scale $s$ 
for any given image coordinate $(u,v)$, thus mapping depth values $z^g \in \D_0^g$ to refined $z^c$:
\begin{equation}
\label{eq:global_z_ref}
\small 
\begin{array}{c}
      z^c = z^g (1 + s) + o \quad \quad \forall z^g \in \D_0^g, \\\\
    \textrm{with }  (o, s) = \Psi(u,v,z^g),
\end{array}
\end{equation}
thus arriving at depth map~$\D_0^c$.

Specifically, $\Psi$ is modeled as a shallow multi-layer perceptron~(MLP):
\begin{equation}
\small 
    (o,s)  = \Psi(\gamma^m(u),\gamma^m(v),\gamma^k(z^g)),
\end{equation}
where $\gamma^m$ and $\gamma^k$ indicate positional encoding~\cite{tancik2020fourfeat} functions of degree $m$ and $k$, respectively.

For optimization of the function, we evaluate $\Psi$ directly at available image positions of the sparse point cloud $\X$ and also compare $\D_0^c$ against $\X$ and use the regularizer as described in Sec.~\ref{sec:loss_reg} using gradient descent.
Evaluating the losses of $\D_0^c$ will prevent overfitting of $\Psi$ solely to the available points.
The weights of $\Psi$ are initialized very small (see the \arxivorelse{Appendix~\ref{suppl:impl_details}}{supplement} for details), thus $o^c$ and $s^c$ will be small, and thus $z^c$ close to $z^g$.

Using a shallow MLP with low frequent positional encoding inherently exploits the spectral bias of MLPs, mapping similar inputs to similar outputs~\cite{tancik2020fourfeat}.
Hence, the re-mapped depth map retains smooth image regions as well as discontinuities.

\subsection{Local Refinement}
\label{sec:local_refinement}
The last refinement stage aims to finetune $\D_0^c$, arriving at $\D_0$. 
Therefore, we jointly optimize depth values $z$ with vertex image coordinates $u,v$ and interject mesh decimation steps in between.
The local refinement step aligns the triangle edges with silhouettes in the image and adjusts depth values to accurately trace the surface.

We introduce the per-vertex inverse scaling factor $f_z$ and image space offsets $\Delta u, \Delta v$, thus allowing for a piece-wise linear approximation of silhouettes, as optimizable parameters:
\begin{equation}
\begin{array}{cc}
    z = \frac{z^c}{f_z},   &  (u,v) = (u^g+\Delta u, v^g+\Delta v).\\
\end{array}
    \label{eq:local_z_ref}
\end{equation}
$\Delta u, \Delta v$ are limited by $\frac{d}{2}$ to ensure that the mesh vertices won't intersect.
(Note that in practice, we enable optimization of $\Delta u, \Delta v$ already during the coarse refinement.) 

As in the previous refinement, $\Delta u,\Delta v$ and $f_z$ are optimized with gradient descent as $\Phi$ is evaluated repeatedly with $(\mathbf{C}_i,\mathbf{R}_i,\mathbf{t}_i)$ and optimized according to the losses described in Sec.~\ref{sec:loss_reg}.

For challenging datasets with varying lighting conditions, we also introduce a tonemapping module that optimizes per image exposure and response parameters. 
For details, we refer the reader to \arxivorelse{Appendix~\ref{suppl:ablation_studies}}{the supplemental material}.

\subsection{Losses \& Regularizers}\label{sec:loss_reg}

We outline losses and regularizers used for the optimization. 
Exact formulas and setups are \arxivorelse{in Appendix~\ref{suppl:loss_reg}}{in the supplemental}.

\paragraph*{Geometric Loss.}
We define $\lossgeo$ as the relative Huber loss~\cite{huber1992robust} between the current depth buffer and the depth values of the projected sparse point cloud.
To get the correspondence between the $\D^*_0$ and $\X$,
we calculate $[\mathbf{R}_0|\mathbf{t}_0]X$ to get the linear $z^x$, and $\mathbf{C}_0[\mathbf{R}_0|\mathbf{t}_0]X$ to get the pixel coordinates per point.
Note that we use the point visibility information that COLMAP provides, so we only consider points that are actually visible in $I_0$. 
We use $\lossgeo$ during coarse and local refinement directly on $\D^*_0$. 
During coarse refinement, we employ $\lossgeo$ on $z^c$ (see Eq.~\ref{eq:global_z_ref}) with $\X$.

\begin{figure}[t]
\centering
\includegraphics[width=1\linewidth]{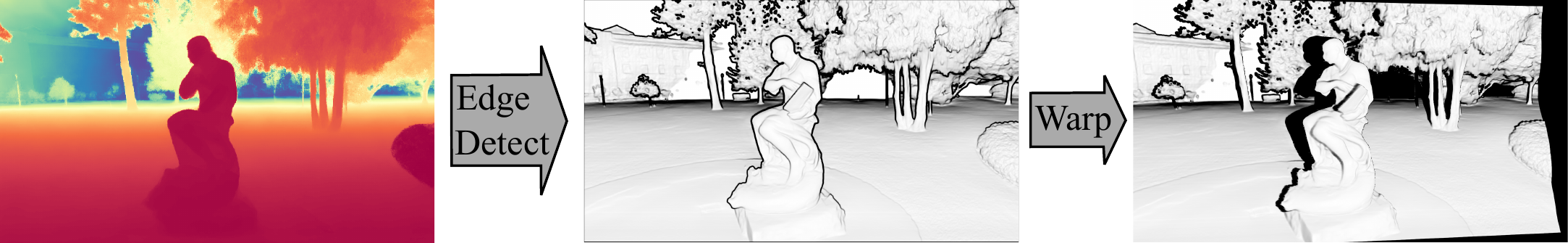}
\begin{tabular}{@{}>{\centering\arraybackslash}p{0.25\linewidth}@{}
            >{\centering\arraybackslash}p{0.12\linewidth}@{}
            >{\centering\arraybackslash}p{0.25\linewidth}@{}
            >{\centering\arraybackslash}p{0.1\linewidth}@{} 
            >{\centering\arraybackslash}p{0.25\linewidth}@{}}
\footnotesize Monocular & &\footnotesize Sobel Filtered & &\footnotesize Projected    \\
\footnotesize  Depth Map $\mathbf{D}_0^r$ & &\footnotesize Edge Map $\mathbf{E}_0$ & &\footnotesize Edge Map $\mathbf{E}_i^*$   \\
\end{tabular}
\vspace{-2mm}
\caption{\label{fig:loss_warped}To ensure proper visibility masking in nearby-view rendering, we project the edge map $\mathbf{E}_0$ onto the rendered view to weight color loss (white fully weighted, black not weighted).
}\vspace{-2mm}

\end{figure}

\paragraph*{Photometric Loss.}
We use a standard mean squared error formulation for our photometric consistency loss $\lossmse$ between rendered $\I^*_i$ and captured images $\I_i$.
However, $\D_0$ cannot represent what is behind visible objects in the scene, so as we render $\I^*_i$, regions that were occluded in $\I_0$ are revealed.
To prevent false comparisons, such regions are masked in the loss.
To this end, we rely on $\D_0^r$ and calculate its edge image~$\E_0$ by utilizing a Sobel filter-based edge detection.
We use the the projected edge map $\E_i^*$ to mask out the loss in occluded regions, which can be seen in Fig.~\ref{fig:loss_warped}.

Optimizing for photometric consistency presents a non-convex challenge with a chance of becoming trapped in local minima if the optimum is too distant. 
Thus, we opted to only apply this loss during local refinement when our current depth estimate $\D^*_0$ is closer to the optimum.

\paragraph*{Regularizers.}
To stabilize the optimization, we apply different regularizers, which carefully balance the contributions of dense photometric consistency, sparse geometric cues and the initial estimation.
Their general idea is to incorporate and reuse the monocular depth image $\D_0^r$ as prior knowledge about structural information and scene layout and apply that to the meshed and \textit{rendered} version $\D_0^*$.

The edge targeting regularizer $\regedge(\E_0^*, \E_0)$, with $\E_0^*$ being the edge image of $\D_0^*$, encourages that silhouettes of the rendered depth map $\D_0^*$ and estimated depth map $\D_0^r$ should align by calculating their mean squared error and an additional term that moves $(u,v)$ on to the edges by minimizing the values sampled from $\E_0$.

A Poisson blending-inspired regularizer $\regpoisson$ is proposed for preserving the local structure from $\D_0^r$ to $\D_0^*$ by assessing their image space gradients.
Again we mask out silhouettes using $\E_0^r$ to neglect discontinuities, which are prone to be inaccurate.
An alternative to reusing the depth map gradients is to regularize on available normals.
Notably, $\regpoisson$ performed on par with using a
SotA normal estimator~\cite{bae2024rethinking} (see \arxivorelse{App.~\ref{suppl:ablation_studies}}{the supplemental material}) and demonstrating that our regularizer should be well applicable also outside of the context of this work.
Our total evaluation loss combines the introduced objective functions using a weighted sum of each of the elements, see \arxivorelse{Appendix~\ref{suppl:impl_details}}{the supplemental material}.

\section{Results}
\label{sec:results}

\subsection{Setup \& Datasets}
We evaluate our method on the synthetic Replica dataset~\cite{straub2019replica, rosinol2023nerf} and the real-world dataset  ScanNet++~\cite{yeshwanth2023scannet++}. 
Both offer dense metric scale ground truth, thus allowing for an isolated evaluation of individual depth maps.

For Replica, we use color and ground truth depth maps from rendered meshes, from 2 room and 2 office scenes randomly selected from the subset prepared by Rosinol et al.~\cite{rosinol2023nerf} and optimizing over 64 views per scene. 
In ScanNet++, ground truth depth maps come from meshes, and the sparse point clouds and poses are computed by COLMAP~\cite{schonberger2016structure}.
For results covered in Sec.~\ref{sec:results_related}, we used 10 ScanNet++ validation scenes, each with 10 depth maps, totaling 100. 
Scenes where chosen to have little blurry images, and only consecutive frames. More details regarding the dataset preparation are available \arxivorelse{in Appendix~\ref{suppl:datasets}}{in the supplement}. 

We evaluate standard error metrics: root mean squared error~(RMSE), mean absolute error~(MAE), L1-rel~\cite{ummenhofer2017demon}, L1-inv~\cite{ummenhofer2017demon}, and $\delta_1$~\cite{yang2024depth}. 
We also assess depth sample accuracy ($|z-z_{GT}| < \tau$) at various thresholds~$\tau$. 
Additionally, we measure the density of depth maps by counting the number of valid pixels.
Values in the top 20 percent are indicated in green, the greener the better.

Notably, we used the exact same hyperparameters for all scenes and datasets, highlighting the stability of our pipeline.
This set of parameters was determined by an extensive set of experiments on withheld ScanNet++ scenes.
Ablation studies regarding the size of the involved mini MLP during coarse refinement, the mesh decimation ratios~$d$ and $r$, 
the tonemapper, different monocular depth estimators~(generally, we used Marigold~\cite{ke2024repurposing}), and multi depthmap refinement can be found in  \arxivorelse{Appendix~\ref{suppl:impl_details} and \ref{suppl:nvdiffrast_support}}{the supplemental material}.
Additionally, we evaluate various loss configurations, even using additional normal estimators, which underlines the flexibility of our method as information from different sources can be trivially fused by adding a loss or adapting the initialization of $\D_0^*$.

\begin{table}[]
\centering
\tiny
\addtolength{\tabcolsep}{-2.5pt}
\begin{tabular}{l|ccc|ccc|c}
 &
  \multicolumn{1}{c}{$\downarrow$\,RMSE} &
  \multicolumn{1}{c}{MAE} &
  L1-inv &
  $\uparrow$\,Acc. &
  \multicolumn{1}{c}{Acc.} &
  \multicolumn{1}{c}{Acc.} &
  Cumul.\\
 &  \multicolumn{1}{c}{} &
  \multicolumn{1}{c}{} &
   &
  $\tau$=0.01\,m &
  \multicolumn{1}{c}{$\tau$=0.05\,m} &
  \multicolumn{1}{c}{$\tau$=0.10\,m} &
  time [s] \\ \hline
Global estimation &
  \cellcolor[HTML]{E67C73}0.29 &
  \cellcolor[HTML]{E67C73}0.21 &
  \cellcolor[HTML]{E67C73}0.17 &
  \cellcolor[HTML]{E67C73}0.06 &
  \cellcolor[HTML]{E67C73}0.27 &
  \cellcolor[HTML]{E67C73}0.47 &
  \\
  (least squares)  & & & & & & & 
  \\\hline
After step: & & & & & & & \\

Init.\,\&\,our global est. ($\mathbf{D}^g_0$) &
  \cellcolor[HTML]{FFFFFF}0.20 &
  \cellcolor[HTML]{FFFFFF}0.14 &
  \cellcolor[HTML]{FFFFFF}0.11 &
  \cellcolor[HTML]{FFFFFF}0.08 &
  \cellcolor[HTML]{FFFFFF}0.32 &
  \cellcolor[HTML]{FFFFFF}0.55 &
 11.2 \\
Meshing ($\M$) &
  \cellcolor[HTML]{FCFEFD}0.20 &
  \cellcolor[HTML]{FFFFFF}0.14 &
  \cellcolor[HTML]{FDFEFE}0.11 &
  \cellcolor[HTML]{FDF9F8}0.08 &
  \cellcolor[HTML]{FDF8F8}0.32 &
  \cellcolor[HTML]{FDF8F8}0.55 &
  13.3\\
Coarse alignment ($\mathbf{D}^c_0$) &
  \cellcolor[HTML]{FAE3E1}0.22 &
  \cellcolor[HTML]{C9E9D9}0.12 &
  \cellcolor[HTML]{FBE8E6}0.12 &
  \cellcolor[HTML]{D1EDDF}0.12 &
  \cellcolor[HTML]{B5E1CB}0.44 &
  \cellcolor[HTML]{A8DCC3}0.66 &
  128\\
Local refinement ($\D_0$) &
  \cellcolor[HTML]{57BB8A}0.17 &
  \cellcolor[HTML]{57BB8A}0.09 &
  \cellcolor[HTML]{57BB8A}0.07 &
  \cellcolor[HTML]{57BB8A}0.22 &
  \cellcolor[HTML]{57BB8A}0.58 &
  \cellcolor[HTML]{57BB8A}0.76 &
  437 \\
\end{tabular}
\caption{Intermediate results. On the ScanNet++ test data, each step of our method significantly improves intermediate outcomes, outperforming the least squares baseline~\cite{ke2024repurposing}. Notably, meshing the depth map preserves quality from $\mathbf{D}^g_0$. Time measured on a Nvidia RTX3090 with OpenGL backend.}
    \label{tab:ablation_every_step}
\end{table}

See Tab.~\ref{tab:ablation_every_step} for a detailed summary of the efficiency each involved step. 
Each step of our method notably enhances results, surpassing the least squares baseline~\cite{ke2024repurposing}.

While this section mainly focuses on per depth map quality, we further evaluated scene global reconstruction quality on a subset of the Tanks and Temples~\cite{knapitschTanksTemplesBenchmarking2017a} dataset,
showing similar results compared to COLMAP.

\subsection{Comparison against Prior Arts}
\label{sec:results_related}

We compare our approach with three highly ranked (and previously best) methods of the Tanks and Temples benchmark~\cite{knapitschTanksTemplesBenchmarking2017a}.
All of them are MVSNet-based methods~\cite{yao2018mvsnet}: MVSFormer~\cite{Cao2022MVSFormerMS}, MVSFormer++~\cite{cao2024mvsformer++}, and GeoMVSNet~\cite{zhang2023geomvsnet}.
For GeoMVSNet however, only weights trained on the DTU dataset were available, which impacted generalization capabilities.
These methods produce a certainty value that can be thresholded to control outlier removal and depth map completeness. 
%
For Replica and ScanNet++ depth map assessments, we focused on valid ground truth depth values within 7 meters, aligning with MVSFormers' bounding volumes, and applied certainty masks at thresholds of 0, 0.25, and 0.5. 
The varied completeness metrics for these depth maps are detailed in the tables.
Additionally, we compare against the recent MASt3R~\cite{leroy2024grounding}, which works on a quite low resolution (512$\times$384) leading to high errors for small details and around edges.
%
As an established representative of patch match-based methods, we also compared with photometric depth maps generated by COLMAP~\cite{schoenberger2016mvs}.
Finally, we compared with the recent indoor-specialized \textit{metric} configuration of the monocular depth estimator DepthAnything2~\cite{yang2024depth2}.
We used the publicly available code (and weights) for comparisons. 

\begin{table}[]
\centering
\tiny
\addtolength{\tabcolsep}{-2.5pt}
\begin{tabular}{l|r|rrr|rrrr}
Method &
  \multicolumn{1}{c|}{$\uparrow$ Valid} & 
  \multicolumn{1}{c}{$\downarrow$\,RMSE} &
  \multicolumn{1}{c}{MAE} &
  \multicolumn{1}{c}{L1-rel} &
  \multicolumn{1}{c}{$\uparrow$ $\delta_1$} &
  \multicolumn{1}{c}{Acc.} &
  \multicolumn{1}{c}{Acc.} &
  \multicolumn{1}{c}{Acc.} \\
  & Samples
   & \multicolumn{1}{c}{}
  &
  &
  &
 &
  \multicolumn{1}{c}{0.01\,m} &
  \multicolumn{1}{c}{0.05\,m} &
  \multicolumn{1}{c}{0.10\,m} \\\hline
MVSFormer (0) &
  \cellcolor[HTML]{58BC8B}1.00 &
  0.80 &
  0.31 &
  0.13 &
  0.88 &
  0.21 &
  0.58 &
  0.73 \\
MVSFormer++ (0) &
  \cellcolor[HTML]{57BB8A}1.00 &
  0.56 &
  0.27 &
  0.10 &
  0.89 &
  0.23 &
  0.61 &
  0.76 \\
MVSFormer (0.25) &
  0.97 &
  0.75 &
  0.27 &
  0.11 &
  0.90 &
  0.21 &
  0.58 &
  0.73 \\
MVSFormer++ (0.25) &
  0.95 &
  0.42 &
  0.17 &
  0.06 &
  0.93 &
  0.23 &
  0.61 &
  0.76 \\
MVSFormer (0.5) &
  0.87 &
  0.46 &
  0.14 &
  0.06 &
  0.96 &
  0.21 &
  0.57 &
  0.71 \\
MVSFormer++ (0.5) &
  0.85 &
  \cellcolor[HTML]{D3EDE0}0.19 &
  \cellcolor[HTML]{CCEADC}0.07 &
  \cellcolor[HTML]{68C296}0.01 &
  0.98 &
  0.22 &
  0.59 &
  0.73 \\
Ours (+) &
  \cellcolor[HTML]{A4DAC0}0.99 &
  \cellcolor[HTML]{57BB8A}0.08 &
  \cellcolor[HTML]{57BB8A}0.04 &
  \cellcolor[HTML]{57BB8A}0.02 &
  \cellcolor[HTML]{57BB8A}1.00 &
  \cellcolor[HTML]{57BB8A}0.37 &
  \cellcolor[HTML]{57BB8A}0.81 &
  \cellcolor[HTML]{57BB8A}0.92 \\
Ours (-) &
  \cellcolor[HTML]{A3DABF}0.99 &
  \cellcolor[HTML]{64C093}0.09 &
  \cellcolor[HTML]{81CCA7}0.05 &
  \cellcolor[HTML]{81CCA7}0.02 &
  \cellcolor[HTML]{64C093}0.99 &
  \cellcolor[HTML]{E4F4EC}0.26 &
  \cellcolor[HTML]{C5E8D6}0.69 &
  \cellcolor[HTML]{95D4B5}0.87
\end{tabular}
\caption{\label{tab:eval_replica}Results on the Replica dataset. In brackets are confidence thresholds for MVSNet-based methods and sparse point cloud accuracy with good (+) and poor (-) quality for our method.}
\end{table}

For the synthetic Replica dataset, sparse point clouds are generated from ground truth data with added realistic levels of noise and outliers. 
Our method excelled the dataset, as noted in Tab.~\ref{tab:eval_replica}, regardless of point cloud quality. 
Here, 'Ours~(+)' denotes high-quality point clouds (2\% noise + 2\% outliers). 
In contrast, 'Ours~(-)'indicates use of very low-quality point clouds (5-10\% noise + 5-10\% outliers) leading to artifacts in extreme cases \arxivorelse{(see Appendix~\ref{suppl:datasets})}{(see supplemental material for visual results and details on data preparation)}.

\begin{table}[]
\tiny
\addtolength{\tabcolsep}{-3pt}
\begin{tabular}{l|r|rrr|rrrr}
Method &
  \multicolumn{1}{c|}{$\uparrow$ Valid} &
  \multicolumn{1}{c}{$\downarrow$\,RMSE} &
  \multicolumn{1}{c}{MAE} &
  \multicolumn{1}{c}{L1-rel} &
  \multicolumn{1}{c}{$\uparrow$ $\delta_1$} &
  \multicolumn{1}{c}{Acc.} &
  \multicolumn{1}{c}{Acc.} &
  \multicolumn{1}{c}{Acc.} \\
  & Samples
   & \multicolumn{1}{c}{}
  &
  &
  &
 &
  \multicolumn{1}{c}{0.01\,m} &
  \multicolumn{1}{c}{0.05\,m} &
  \multicolumn{1}{c}{0.10\,m} \\\hline
DepthAnything2 & & & & & & & \\
(metric) & \cellcolor[HTML]{57BB8A}100\% & 1.91 & \cellcolor[HTML]{FFFFFF}1.79 & \cellcolor[HTML]{FFFFFF}1.51 & 0.00 & 0.00 & 0.00 & 0.00 \\
COLMAP                  & \cellcolor[HTML]{FFFFFF}69\%  & 0.60 & \cellcolor[HTML]{BDE4D1}0.30 & \cellcolor[HTML]{BEE5D2}0.24 & 0.73 & 0.17 & 0.35 & 0.41 \\
MASt3R &
  \cellcolor[HTML]{6AC397}98\% &
  \cellcolor[HTML]{65C093}0.19 &
  \cellcolor[HTML]{78C8A1}0.16 &
  \cellcolor[HTML]{8DD0AF}0.16 &
  \cellcolor[HTML]{91D3B3}0.89 &
  0.07 &
  0.30 &
  0.51 \\
MVSFormer$_{0}$         & \cellcolor[HTML]{58BC8B}100\% & 1.08 & \cellcolor[HTML]{FFFFFF}0.47 & \cellcolor[HTML]{FFFFFF}0.41 & 0.68 & 0.16 & 0.40 & 0.52 \\
MVSFormer++$_{0}$ &
  \cellcolor[HTML]{58BC8B}100\% &
  0.67 &
  \cellcolor[HTML]{B3E0CA}0.28 &
  \cellcolor[HTML]{BBE3CF}0.24 &
  \cellcolor[HTML]{EEF8F3}0.79 &
  \cellcolor[HTML]{57BB8A}0.22 &
  \cellcolor[HTML]{ABDDC5}0.53 &
  \cellcolor[HTML]{D0ECDF}0.65 \\
GeoMVSNet$_{0}$         & \cellcolor[HTML]{57BB8A}100\% & 1.19 & \cellcolor[HTML]{FFFFFF}0.61 & \cellcolor[HTML]{FFFFFF}0.77 & 0.53 & 0.05 & 0.18 & 0.28 \\
MVSFormer$_{0.25}$      & \cellcolor[HTML]{C2E7D5}90\%  & 1.09 & \cellcolor[HTML]{FFFFFF}0.46 & \cellcolor[HTML]{FFFFFF}0.38 & 0.72 & 0.16 & 0.40 & 0.51 \\
MVSFormer++$_{0.25}$ &
  \cellcolor[HTML]{DBF1E6}88\% &
  0.61 &
  \cellcolor[HTML]{99D5B8}0.22 &
  \cellcolor[HTML]{9DD7BB}0.19 &
  \cellcolor[HTML]{ADDEC6}0.86 &
  \cellcolor[HTML]{76C8A0}0.22 &
  \cellcolor[HTML]{BAE3CF}0.52 &
  \cellcolor[HTML]{E5F5ED}0.63 \\
GeoMVSNet$_{0.25}$      & \cellcolor[HTML]{58BC8B}100\% & 1.19 & \cellcolor[HTML]{FFFFFF}0.61 & \cellcolor[HTML]{FFFFFF}0.77 & 0.53 & 0.05 & 0.18 & 0.28 \\
MVSFormer$_{0.5}$ &
  \cellcolor[HTML]{FFFFFF}47\% &
  0.52 &
  \cellcolor[HTML]{79C9A2}0.16 &
  \cellcolor[HTML]{69C296}0.10 &
  \cellcolor[HTML]{72C69D}0.93 &
  0.14 &
  0.30 &
  0.36 \\
MVSFormer++$_{0.5}$ &
  \cellcolor[HTML]{FFFFFF}62\% &
  \cellcolor[HTML]{85CEAA}0.26 &
  \cellcolor[HTML]{5BBC8C}0.10 &
  \cellcolor[HTML]{57BB8A}0.07 &
  \cellcolor[HTML]{57BB8A}0.96 &
  0.19 &
  0.43 &
  0.50 \\
GeoMVSNet$_{0.5}$       & \cellcolor[HTML]{FFFFFF}23\%  & 1.45 & \cellcolor[HTML]{FFFFFF}0.90 & \cellcolor[HTML]{FFFFFF}1.11 & 0.38 & 0.00 & 0.02 & 0.04 \\
Ours &
  \cellcolor[HTML]{5DBE8E}99\% &
  \cellcolor[HTML]{57BB8A}0.17 &
  \cellcolor[HTML]{57BB8A}0.090 &
  \cellcolor[HTML]{5BBC8D}0.07 &
  \cellcolor[HTML]{71C69D}0.93 &
  \cellcolor[HTML]{6EC59A}0.22 &
  \cellcolor[HTML]{57BB8A}0.58 &
  \cellcolor[HTML]{57BB8A}0.76
\end{tabular}\vspace{-2mm}
\caption{\label{tab:eval_scannetpp}Results on the ScanNet++ dataset~\cite{yeshwanth2023scannet++}. Numbers in subscript indicate the confidence threshold for MVSNet methods.}
\end{table}

Quantitative results for ScanNet++ are presented in Tab.~\ref{tab:eval_scannetpp}. 
We achieved top or on-par results in RMSE, completeness (valid samples), MAE, L1-rel, and accuracy at $\tau=0.05$ and $\tau=0.1$. 
MVSFormer++ with a certainty threshold of 0.5 slightly outperformed us in L1-inv, $\delta_1$, and accuracy at $\tau=0.01$. 
However, their results only included 62\% of values, compared to our over 99.5\% of samples.

The qualitative results and ScanNet++ are shown in Fig.~\ref{fig:eval_scannet}. 
Our outputs are clean, dense, and plausible. 
However, some artifacts appear on the glossy table in \texttt{Scene 3}. 
Sometimes, our results surpass the ground truth, as in the crop of \texttt{Scene 4} or the table leg in \texttt{Scene 5}, right crop. 
MVSNet-like results are sparse with noisy walls. 
DepthAnything2's metric version offers promising details, though its global scale is inaccurate.

\arxivorelse{
\begin{figure*}
    \centering
    \includegraphics[width=0.91\linewidth]{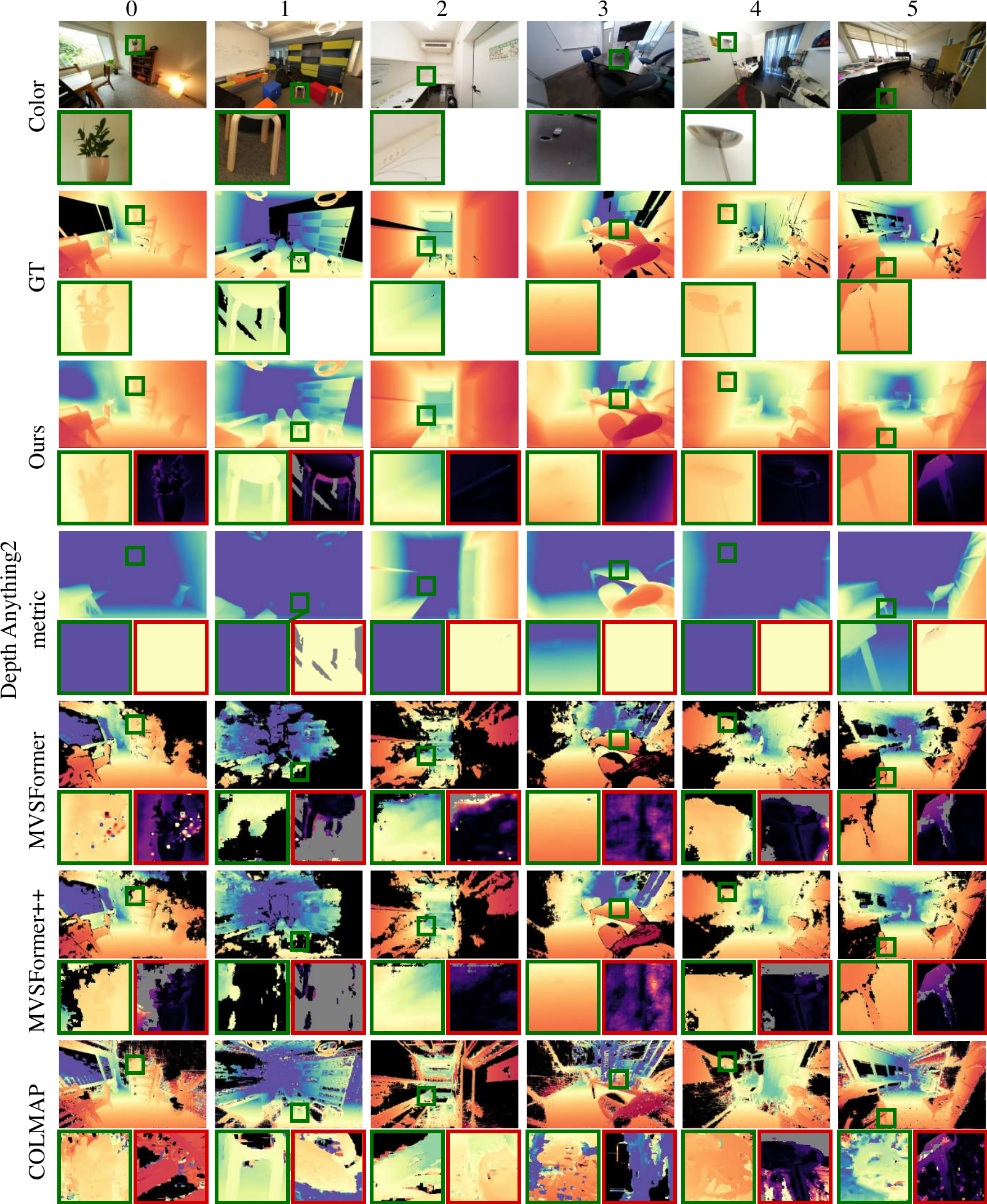}
    \caption{Qualitative results on the ScanNet++ dataset, with  MVSFormers' certainty threshold at their default of 0.5. Green borders indicate image crops, while red highlights error maps; brighter areas imply greater errors, and grey indicates missing samples.}
    \label{fig:eval_scannet}
\end{figure*}
}{
\begin{figure}
    \centering
    \includegraphics[width=0.98\linewidth]{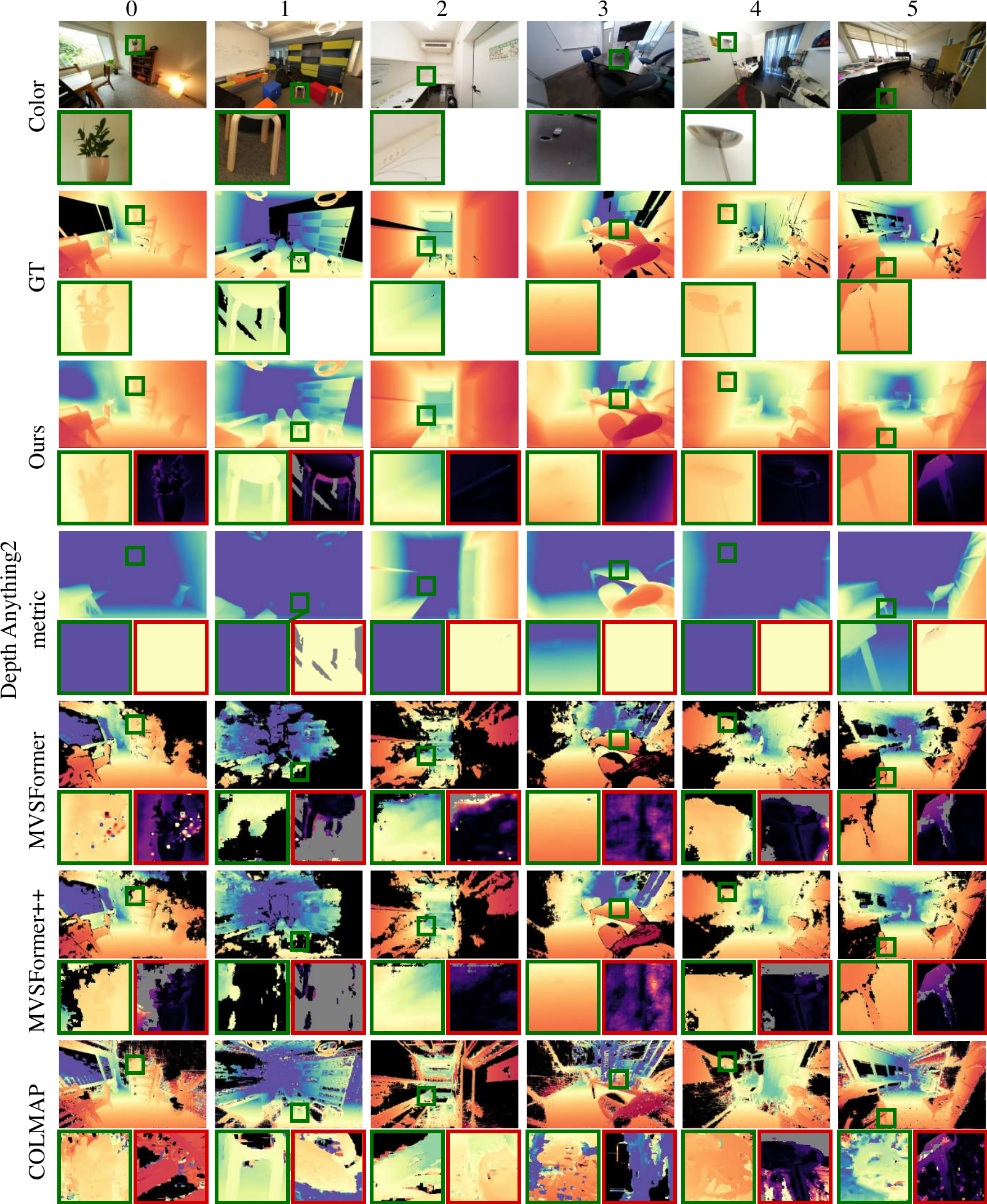}\vspace{-1mm}
    \caption{Qualitative results on the ScanNet++ dataset, with  MVSFormers' certainty threshold at their default of 0.5. Green borders indicate image crops, while red highlights error maps; brighter areas imply greater errors, and grey indicates missing samples.}
    \label{fig:eval_scannet}
\pagebreak

\end{figure}
}

\section{Limitations}
While our method shows strong performance, it still has limitations. 
Photometric consistency proves useful, but fails in cases with specular materials (see e.g. Fig.~\ref{fig:eval_scannet}) which our method does not account for. 
This could be addressed with uncertainty estimation, anchoring optimization with high-certainty depth estimates.

Ultimately, our approach relies on the initial input. 
Outliers in the sparse point cloud can result in spurious artifacts and the scene topology, in form of an edge image, has to be derived from the monocular depth map. 
The edge image is crucial for accurately masking occluded areas, influencing effectiveness of the losses. 

\section{Conclusion}

In conclusion, our approach demonstrates that monocular depth maps can be significantly improved by incorporating photometric consistency across multiple views, leading to enhanced depth estimation. 
By optimizing a meshed version of the monocular depth map with a differentiable renderer, we directly leverage the strong initial estimates and achieve stable optimization, which refines scaling and corrects errors. 
In this context, our Poisson and edge regularizers have proven to be particularly effective, ensuring robust and reliable depth map refinement.
Our simple yet effective pipeline consistently produces accurate and dense depth maps, which is highly desirable for downstream tasks.

\section{Acknowledgments}

We would like to thank all members of the Visual Computing Lab Erlangen for the fruitful discussions. 
Specifically, we appreciate Darius Rückert's support during late night debugging sessions.
The authors gratefully acknowledge the scientific support and HPC resources provided by the National High Performance Computing Center of the Friedrich-Alexander-Universität Erlangen-Nürnberg (NHR@FAU) under the project b212dc. 
NHR funding is provided by federal and Bavarian state authorities. NHR@FAU hardware is partially funded by the German Research Foundation (DFG) – 440719683. 
Linus Franke was supported by the 5G innovation program of the German Federal Ministry for Digital and Transport under the funding code 165GU103B. 
Joachim Keinert was supported by the Free State of Bavaria in the DSAI project.

{
    \small
    \bibliographystyle{ieeenat_fullname}
    \bibliography{egbib}

\begin{thebibliography}{79}
\providecommand{\natexlab}[1]{#1}
\providecommand{\url}[1]{\texttt{#1}}
\expandafter\ifx\csname urlstyle\endcsname\relax
  \providecommand{\doi}[1]{doi: #1}\else
  \providecommand{\doi}{doi: \begingroup \urlstyle{rm}\Url}\fi

\bibitem[Bae and Davison(2024)]{bae2024rethinking}
Gwangbin Bae and Andrew~J Davison.
\newblock Rethinking inductive biases for surface normal estimation.
\newblock In \emph{Proceedings of the IEEE/CVF Conference on Computer Vision
  and Pattern Recognition}, pages 9535--9545, 2024.

\bibitem[Bleyer et~al.(2011)Bleyer, Rhemann, and Rother]{bleyer2011patchmatch}
Michael Bleyer, Christoph Rhemann, and Carsten Rother.
\newblock Patchmatch stereo-stereo matching with slanted support windows.
\newblock In \emph{BMVC}, pages 1--11, 2011.

\bibitem[Buss(2003)]{buss20033d}
Samuel~R Buss.
\newblock \emph{3D computer graphics: a mathematical introduction with OpenGL}.
\newblock Cambridge University Press, 2003.

\bibitem[Cao et~al.(2022)Cao, Ren, and Fu]{Cao2022MVSFormerMS}
Chenjie Cao, Xinlin Ren, and Yanwei Fu.
\newblock {{MVSFormer}}: {{Multi-view}} stereo by learning robust image
  features and temperature-based depth.
\newblock \emph{Trans. Mach. Learn. Res.}, 2022, 2022.

\bibitem[Cao et~al.(2024)Cao, Ren, and Yanwei]{cao2024mvsformer++}
Chenjie Cao, Xinlin Ren, and Fu Yanwei.
\newblock {{MVSFormer}}++: {{Revealing}} the devil in transformer's details for
  multi-view stereo.
\newblock In \emph{International Conference on Learning Representations
  ({{ICLR}})}, 2024.

\bibitem[Cheng et~al.(2020)Cheng, Xu, Zhu, Li, Li, Ramamoorthi, and
  Su]{cheng2020deep}
Shuo Cheng, Zexiang Xu, Shilin Zhu, Zhuwen Li, Li~Erran Li, Ravi Ramamoorthi,
  and Hao Su.
\newblock Deep stereo using adaptive thin volume representation with
  uncertainty awareness.
\newblock In \emph{Proceedings of the IEEE/CVF Conference on Computer Vision
  and Pattern Recognition}, pages 2524--2534, 2020.

\bibitem[Chugunov et~al.(2023)Chugunov, Zhang, and Heide]{chugunov2023shakes}
Ilya Chugunov, Yuxuan Zhang, and Felix Heide.
\newblock Shakes on a plane: {{Unsupervised}} depth estimation from
  unstabilized photography.
\newblock In \emph{Proceedings of the {{IEEE}}/{{CVF}} Conference on Computer
  Vision and Pattern Recognition}, pages 13240--13251, 2023.

\bibitem[Collins(1996)]{collins1996space}
Robert~T Collins.
\newblock A space-sweep approach to true multi-image matching.
\newblock In \emph{Proceedings {{CVPR IEEE}} Computer Society Conference on
  Computer Vision and Pattern Recognition}, pages 358--363. IEEE, 1996.

\bibitem[Ding et~al.(2022)Ding, Yuan, Zhu, Zhang, Liu, Wang, and
  Liu]{ding2022transmvsnet}
Yikang Ding, Wentao Yuan, Qingtian Zhu, Haotian Zhang, Xiangyue Liu, Yuanjiang
  Wang, and Xiao Liu.
\newblock Transmvsnet: Global context-aware multi-view stereo network with
  transformers.
\newblock In \emph{Proceedings of the IEEE/CVF Conference on Computer Vision
  and Pattern Recognition}, pages 8585--8594, 2022.

\bibitem[Dong et~al.(2022)Dong, Garratt, Anavatti, and Abbass]{dong2022towards}
Xingshuai Dong, Matthew~A Garratt, Sreenatha~G Anavatti, and Hussein~A Abbass.
\newblock Towards real-time monocular depth estimation for robotics: A survey.
\newblock \emph{IEEE Transactions on Intelligent Transportation Systems},
  23\penalty0 (10):\penalty0 16940--16961, 2022.

\bibitem[Fink et~al.(2023)Fink, R{\"u}ckert, Franke, Keinert, and
  Stamminger]{fink2023livenvs}
Laura Fink, Darius R{\"u}ckert, Linus Franke, Joachim Keinert, and Marc
  Stamminger.
\newblock Livenvs: {{Neural}} view synthesis on live rgb-d streams.
\newblock In \emph{{{SIGGRAPH}} Asia 2023 Conference Papers}, pages 1--11,
  2023.

\bibitem[Franke et~al.(2024)Franke, Rückert, Fink, and
  Stamminger]{franke2024trips}
Linus Franke, Darius Rückert, Laura Fink, and Marc Stamminger.
\newblock {TRIPS: Trilinear Point Splatting for Real-Time Radiance Field
  Rendering}.
\newblock \emph{Computer Graphics Forum}, 43\penalty0 (2):\penalty0 e15012,
  2024.

\bibitem[Garland and Heckbert(1997)]{garland1997surface}
Michael Garland and Paul~S Heckbert.
\newblock Surface simplification using quadric error metrics.
\newblock In \emph{Proceedings of the 24th Annual Conference on {{Computer}}
  Graphics and Interactive Techniques}, pages 209--216, 1997.

\bibitem[Ge et~al.(2024)Ge, Xu, Zhao, Sun, Huang, Sun, Chen, and
  Shen]{ge2024geobench}
Yongtao Ge, Guangkai Xu, Zhiyue Zhao, Libo Sun, Zheng Huang, Yanlong Sun, Hao
  Chen, and Chunhua Shen.
\newblock {{GeoBench}}: {{Benchmarking}} and analyzing monocular geometry
  estimation models.
\newblock \emph{arXiv preprint arXiv:2406.12671}, 2024.

\bibitem[Gruber et~al.(2019)Gruber, {Julca-Aguilar}, Bijelic, and
  Heide]{gruber2019gated2depth}
Tobias Gruber, Frank {Julca-Aguilar}, Mario Bijelic, and Felix Heide.
\newblock Gated2depth: {{Real-time}} dense lidar from gated images.
\newblock In \emph{Proceedings of the {{IEEE}}/{{CVF}} International Conference
  on Computer Vision}, pages 1506--1516, 2019.

\bibitem[Gu et~al.(2020)Gu, Fan, Zhu, Dai, Tan, and Tan]{gu2020cascade}
Xiaodong Gu, Zhiwen Fan, Siyu Zhu, Zuozhuo Dai, Feitong Tan, and Ping Tan.
\newblock Cascade cost volume for high-resolution multi-view stereo and stereo
  matching.
\newblock In \emph{Proceedings of the IEEE/CVF conference on computer vision
  and pattern recognition}, pages 2495--2504, 2020.

\bibitem[Haefner et~al.(2018)Haefner, Qu{\'e}au, M{\"o}llenhoff, and
  Cremers]{haefner2018fight}
Bjoern Haefner, Yvain Qu{\'e}au, Thomas M{\"o}llenhoff, and Daniel Cremers.
\newblock Fight ill-posedness with ill-posedness: {{Single-shot}} variational
  depth super-resolution from shading.
\newblock In \emph{Proceedings of the {{IEEE}} Conference on Computer Vision
  and Pattern Recognition}, pages 164--174, 2018.

\bibitem[Hahlbohm et~al.(2025)Hahlbohm, Franke, Kappel, Castillo, Eisemann,
  Stamminger, and Magnor]{hahlbohm2024inpc}
Florian Hahlbohm, Linus Franke, Moritz Kappel, Susana Castillo, Martin
  Eisemann, Marc Stamminger, and Marcus Magnor.
\newblock Inpc: Implicit neural point clouds for radiance field rendering.
\newblock In \emph{2025 International Conference on 3D Vision (3DV)}, 2025.

\bibitem[Hartley and Zisserman(2003)]{hartley2003multiple}
Richard Hartley and Andrew Zisserman.
\newblock \emph{Multiple View Geometry in Computer Vision}.
\newblock Cambridge university press, 2003.

\bibitem[Hedman et~al.(2018)Hedman, Philip, Price, Frahm, Drettakis, and
  Brostow]{hedman2018deep}
Peter Hedman, Julien Philip, True Price, Jan-Michael Frahm, George Drettakis,
  and Gabriel Brostow.
\newblock Deep blending for free-viewpoint image-based rendering.
\newblock \emph{ACM Transactions on Graphics (ToG)}, 37\penalty0 (6):\penalty0
  1--15, 2018.

\bibitem[Huber(1992)]{huber1992robust}
Peter~J Huber.
\newblock Robust estimation of a location parameter.
\newblock In \emph{Breakthroughs in Statistics: {{Methodology}} and
  Distribution}, pages 492--518. Springer, 1992.

\bibitem[Ito et~al.(2024)Ito, Miura, Ito, and Aoki]{ito2024neural}
Shintaro Ito, Kanta Miura, Koichi Ito, and Takafumi Aoki.
\newblock Neural radiance field-inspired depth map refinement for accurate
  multi-view stereo.
\newblock \emph{Journal of Imaging}, 10\penalty0 (3):\penalty0 68, 2024.

\bibitem[Kadambi et~al.(2017)Kadambi, Taamazyan, Shi, and
  Raskar]{kadambi2017depth}
Achuta Kadambi, Vage Taamazyan, Boxin Shi, and Ramesh Raskar.
\newblock Depth sensing using geometrically constrained polarization normals.
\newblock \emph{International Journal of Computer Vision}, 125:\penalty0
  34--51, 2017.

\bibitem[Kalia et~al.(2019)Kalia, Navab, and Salcudean]{kalia2019real}
Megha Kalia, Nassir Navab, and Tim Salcudean.
\newblock A real-time interactive augmented reality depth estimation technique
  for surgical robotics.
\newblock In \emph{2019 international conference on robotics and automation
  (icra)}, pages 8291--8297. IEEE, 2019.

\bibitem[Ke et~al.(2024)Ke, Obukhov, Huang, Metzger, Daudt, and
  Schindler]{ke2024repurposing}
Bingxin Ke, Anton Obukhov, Shengyu Huang, Nando Metzger, Rodrigo~Caye Daudt,
  and Konrad Schindler.
\newblock Repurposing diffusion-based image generators for monocular depth
  estimation.
\newblock In \emph{Proceedings of the {{IEEE}}/{{CVF}} Conference on Computer
  Vision and Pattern Recognition}, pages 9492--9502, 2024.

\bibitem[Kim et~al.()Kim, Park, Yu, Sohn, and Shin]{kimmc2}
Subin Kim, Seong~Hyeon Park, Sihyun Yu, Kihyuk Sohn, and Jinwoo Shin.
\newblock {{MC2}}: {{Multi-view}} consistent depth estimation via coordinated
  image-based neural rendering.

\bibitem[Kingma(2014)]{kingma2014adam}
{\relax DP} Kingma.
\newblock Adam: A method for stochastic optimization.
\newblock \emph{arXiv preprint arXiv:1412.6980}, 2014.

\bibitem[Knapitsch et~al.(2017)Knapitsch, Park, Zhou, and
  Koltun]{knapitschTanksTemplesBenchmarking2017a}
Arno Knapitsch, Jaesik Park, Qian-Yi Zhou, and Vladlen Koltun.
\newblock Tanks and {{Temples}}: {{Benchmarking Large-Scale Scene
  Reconstruction}}.
\newblock \emph{ACM Transactions on Graphics}, 36\penalty0 (4), 2017.

\bibitem[Laga et~al.(2020)Laga, Jospin, Boussaid, and
  Bennamoun]{laga2020survey}
Hamid Laga, Laurent~Valentin Jospin, Farid Boussaid, and Mohammed Bennamoun.
\newblock A survey on deep learning techniques for stereo-based depth
  estimation.
\newblock \emph{IEEE Transactions on Pattern Analysis and Machine
  Intelligence}, 44\penalty0 (4):\penalty0 1738--1764, 2020.

\bibitem[Laine et~al.(2020)Laine, Hellsten, Karras, Seol, Lehtinen, and
  Aila]{Laine2020diffrast}
Samuli Laine, Janne Hellsten, Tero Karras, Yeongho Seol, Jaakko Lehtinen, and
  Timo Aila.
\newblock Modular primitives for high-performance differentiable rendering.
\newblock \emph{ACM Transactions on Graphics (ToG)}, 39\penalty0 (6), 2020.

\bibitem[Leroy et~al.(2024)Leroy, Cabon, and Revaud]{leroy2024grounding}
Vincent Leroy, Yohann Cabon, and J{\'e}r{\^o}me Revaud.
\newblock Grounding image matching in 3d with mast3r.
\newblock In \emph{European Conference on Computer Vision}, pages 71--91.
  Springer, 2024.

\bibitem[Li et~al.(2024)Li, Zhang, Bai, Zheng, Ning, Zhou, and
  Gu]{li2024dngaussian}
Jiahe Li, Jiawei Zhang, Xiao Bai, Jin Zheng, Xin Ning, Jun Zhou, and Lin Gu.
\newblock Dngaussian: {{Optimizing}} sparse-view 3d gaussian radiance fields
  with global-local depth normalization.
\newblock In \emph{Proceedings of the {{IEEE}}/{{CVF}} Conference on Computer
  Vision and Pattern Recognition}, pages 20775--20785, 2024.

\bibitem[Li et~al.(2023)Li, Wang, Cole, Tucker, and Snavely]{li2023dynibar}
Zhengqi Li, Qianqian Wang, Forrester Cole, Richard Tucker, and Noah Snavely.
\newblock Dynibar: {{Neural}} dynamic image-based rendering.
\newblock In \emph{Proceedings of the {{IEEE}}/{{CVF}} Conference on Computer
  Vision and Pattern Recognition}, pages 4273--4284, 2023.

\bibitem[Liao et~al.(2022)Liao, Ding, Shavit, Huang, Ren, Guo, Feng, and
  Zhang]{liao2022wt}
Jinli Liao, Yikang Ding, Yoli Shavit, Dihe Huang, Shihao Ren, Jia Guo, Wensen
  Feng, and Kai Zhang.
\newblock Wt-mvsnet: window-based transformers for multi-view stereo.
\newblock \emph{Advances in Neural Information Processing Systems},
  35:\penalty0 8564--8576, 2022.

\bibitem[Liu et~al.(2023)Liu, Ye, Zhao, Pan, Shi, and Cao]{liu2023epipolar}
Tianqi Liu, Xinyi Ye, Weiyue Zhao, Zhiyu Pan, Min Shi, and Zhiguo Cao.
\newblock When epipolar constraint meets non-local operators in multi-view
  stereo.
\newblock In \emph{Proceedings of the IEEE/CVF International Conference on
  Computer Vision}, pages 18088--18097, 2023.

\bibitem[Ma et~al.(2019)Ma, Cavalheiro, and Karaman]{ma2019self}
Fangchang Ma, Guilherme~Venturelli Cavalheiro, and Sertac Karaman.
\newblock Self-supervised sparse-to-dense: {{Self-supervised}} depth completion
  from lidar and monocular camera.
\newblock In \emph{2019 International Conference on Robotics and Automation
  ({{ICRA}})}, pages 3288--3295. IEEE, 2019.

\bibitem[Mirdehghan et~al.()Mirdehghan, Wu, Chen, Lindell, and
  Kutulakos]{mirdehghanTurboSLDenseAccurate}
Parsa Mirdehghan, Maxx Wu, Wenzheng Chen, David~B Lindell, and Kiriakos~N
  Kutulakos.
\newblock {{TurboSL}}: {{Dense Accurate}} and {{Fast 3D}} by {{Neural Inverse
  Structured Light}}.
\newblock \emph{Proceedings {{CVPR IEEE}} Computer Society Conference on
  Computer Vision and Pattern Recognition}.

\bibitem[M{\"u}ller et~al.(2022)M{\"u}ller, Evans, Schied, and
  Keller]{muller2022instant}
Thomas M{\"u}ller, Alex Evans, Christoph Schied, and Alexander Keller.
\newblock Instant neural graphics primitives with a multiresolution hash
  encoding.
\newblock \emph{ACM Transactions on Graphics (TOG)}, 41\penalty0 (4):\penalty0
  1--15, 2022.

\bibitem[{Or-El} et~al.(2015){Or-El}, Rosman, Wetzler, Kimmel, and
  Bruckstein]{or2015rgbd}
Roy {Or-El}, Guy Rosman, Aaron Wetzler, Ron Kimmel, and Alfred~M Bruckstein.
\newblock Rgbd-fusion: {{Real-time}} high precision depth recovery.
\newblock In \emph{Proceedings of the {{IEEE}} Conference on Computer Vision
  and Pattern Recognition}, pages 5407--5416, 2015.

\bibitem[{Or-El} et~al.(2016){Or-El}, Hershkovitz, Wetzler, Rosman, Bruckstein,
  and Kimmel]{or2016real}
Roy {Or-El}, Rom Hershkovitz, Aaron Wetzler, Guy Rosman, Alfred~M Bruckstein,
  and Ron Kimmel.
\newblock Real-time depth refinement for specular objects.
\newblock In \emph{Proceedings of the {{IEEE}} Conference on Computer Vision
  and Pattern Recognition}, pages 4378--4386, 2016.

\bibitem[{\"O}rnek et~al.(2022){\"O}rnek, Mudgal, Wald, Wang, Navab, and
  Tombari]{ornek20222d}
Evin~P{\i}nar {\"O}rnek, Shristi Mudgal, Johanna Wald, Yida Wang, Nassir Navab,
  and Federico Tombari.
\newblock From 2d to 3d: Re-thinking benchmarking of monocular depth
  prediction.
\newblock \emph{arXiv preprint arXiv:2203.08122}, 2022.

\bibitem[Park et~al.(2024)Park, Li, and Kitani]{park2024flexible}
Jinhyung Park, Yu-Jhe Li, and Kris Kitani.
\newblock Flexible depth completion for sparse and varying point densities.
\newblock In \emph{Proceedings of the IEEE/CVF Conference on Computer Vision
  and Pattern Recognition}, pages 21540--21550, 2024.

\bibitem[P{\'e}rez et~al.(2003)P{\'e}rez, Gangnet, and Blake]{perez2003poisson}
Patrick P{\'e}rez, Michel Gangnet, and Andrew Blake.
\newblock Poisson image editing.
\newblock In \emph{{{ACM SIGGRAPH}} 2003 Papers}, pages 313--318. 2003.

\bibitem[Ranftl et~al.(2020)Ranftl, Lasinger, Hafner, Schindler, and
  Koltun]{ranftl2020towards}
Ren{\'e} Ranftl, Katrin Lasinger, David Hafner, Konrad Schindler, and Vladlen
  Koltun.
\newblock Towards robust monocular depth estimation: Mixing datasets for
  zero-shot cross-dataset transfer.
\newblock \emph{IEEE Transactions on Pattern Analysis and Machine
  Intelligence}, 44\penalty0 (3):\penalty0 1623--1637, 2020.

\bibitem[Rosinol et~al.(2023)Rosinol, Leonard, and Carlone]{rosinol2023nerf}
Antoni Rosinol, John~J Leonard, and Luca Carlone.
\newblock Nerf-slam: {{Real-time}} dense monocular slam with neural radiance
  fields.
\newblock In \emph{2023 {{IEEE}}/{{RSJ}} International Conference on
  Intelligent Robots and Systems ({{IROS}})}, pages 3437--3444. IEEE, 2023.

\bibitem[R{\"u}ckert et~al.(2022)R{\"u}ckert, Franke, and
  Stamminger]{ruckert2022adop}
Darius R{\"u}ckert, Linus Franke, and Marc Stamminger.
\newblock {ADOP: Approximate Differentiable One-Pixel Point Rendering}.
\newblock \emph{ACM Trans. Graph.}, 41\penalty0 (4), 2022.

\bibitem[Schonberger and Frahm(2016)]{schonberger2016structure}
Johannes~L Schonberger and Jan-Michael Frahm.
\newblock Structure-from-motion revisited.
\newblock In \emph{Proceedings of the {{IEEE}} Conference on Computer Vision
  and Pattern Recognition}, pages 4104--4113, 2016.

\bibitem[Sch{\"o}nberger et~al.(2016)Sch{\"o}nberger, Zheng, Pollefeys, and
  Frahm]{schoenberger2016mvs}
Johannes~Lutz Sch{\"o}nberger, Enliang Zheng, Marc Pollefeys, and Jan-Michael
  Frahm.
\newblock Pixelwise view selection for unstructured multi-view stereo.
\newblock In \emph{European Conference on Computer Vision ({{ECCV}})}, 2016.

\bibitem[Shandilya et~al.(2023)Shandilya, Attal, Richardt, Tompkin, and
  O'toole]{shandilya2023neural}
Aarrushi Shandilya, Benjamin Attal, Christian Richardt, James Tompkin, and
  Matthew O'toole.
\newblock Neural fields for structured lighting.
\newblock In \emph{Proceedings of the {{IEEE}}/{{CVF}} International Conference
  on Computer Vision}, pages 3512--3522, 2023.

\bibitem[Spencer et~al.(2024)Spencer, Tosi, Poggi, Arora, Russell, Hadfield,
  Bowden, Zhou, Li, Rao, et~al.]{spencer2024third}
Jaime Spencer, Fabio Tosi, Matteo Poggi, Ripudaman~Singh Arora, Chris Russell,
  Simon Hadfield, Richard Bowden, GuangYuan Zhou, ZhengXin Li, Qiang Rao,
  et~al.
\newblock The third monocular depth estimation challenge.
\newblock In \emph{Proceedings of the IEEE/CVF Conference on Computer Vision
  and Pattern Recognition}, pages 1--14, 2024.

\bibitem[Sterzentsenko et~al.(2019)Sterzentsenko, Saroglou, Chatzitofis,
  Thermos, Zioulis, Doumanoglou, Zarpalas, and Daras]{sterzentsenko2019self}
Vladimiros Sterzentsenko, Leonidas Saroglou, Anargyros Chatzitofis, Spyridon
  Thermos, Nikolaos Zioulis, Alexandros Doumanoglou, Dimitrios Zarpalas, and
  Petros Daras.
\newblock Self-supervised deep depth denoising.
\newblock In \emph{Proceedings of the {{IEEE}}/{{CVF}} International Conference
  on Computer Vision}, pages 1242--1251, 2019.

\bibitem[Stevens et~al.(2020)Stevens, Antiga, and Viehmann]{stevens2020deep}
Eli Stevens, Luca Antiga, and Thomas Viehmann.
\newblock \emph{Deep Learning with {{PyTorch}}}.
\newblock Manning Publications, 2020.

\bibitem[Straub et~al.(2019)Straub, Whelan, Ma, Chen, Wijmans, Green, Engel,
  {Mur-Artal}, Ren, Verma, et~al.]{straub2019replica}
Julian Straub, Thomas Whelan, Lingni Ma, Yufan Chen, Erik Wijmans, Simon Green,
  Jakob~J Engel, Raul {Mur-Artal}, Carl Ren, Shobhit Verma, et~al.
\newblock The {{Replica}} dataset: {{A}} digital replica of indoor spaces.
\newblock \emph{arXiv preprint arXiv:1906.05797}, 2019.

\bibitem[Sun et~al.(2023)Sun, Ye, Xiong, Choe, Wang, Su, and
  Ranjan]{sun2023consistent}
Zhanghao Sun, Wei Ye, Jinhui Xiong, Gyeongmin Choe, Jialiang Wang, Shuochen Su,
  and Rakesh Ranjan.
\newblock Consistent direct time-of-flight video depth super-resolution.
\newblock In \emph{Proceedings of the Ieee/Cvf Conference on Computer Vision
  and Pattern Recognition}, pages 5075--5085, 2023.

\bibitem[Tancik et~al.(2020)Tancik, Srinivasan, Mildenhall, {Fridovich-Keil},
  Raghavan, Singhal, Ramamoorthi, Barron, and Ng]{tancik2020fourfeat}
Matthew Tancik, Pratul~P. Srinivasan, Ben Mildenhall, Sara {Fridovich-Keil},
  Nithin Raghavan, Utkarsh Singhal, Ravi Ramamoorthi, Jonathan~T. Barron, and
  Ren Ng.
\newblock Fourier features let networks learn high frequency functions in low
  dimensional domains.
\newblock \emph{NeurIPS}, 2020.

\bibitem[Tao et~al.(2015)Tao, Srinivasan, Malik, Rusinkiewicz, and
  Ramamoorthi]{tao2015depth}
Michael~W Tao, Pratul~P Srinivasan, Jitendra Malik, Szymon Rusinkiewicz, and
  Ravi Ramamoorthi.
\newblock Depth from shading, defocus, and correspondence using light-field
  angular coherence.
\newblock In \emph{Proceedings of the {{IEEE}} Conference on Computer Vision
  and Pattern Recognition}, pages 1940--1948, 2015.

\bibitem[Ummenhofer et~al.(2017)Ummenhofer, Zhou, Uhrig, Mayer, Ilg,
  Dosovitskiy, and Brox]{ummenhofer2017demon}
Benjamin Ummenhofer, Huizhong Zhou, Jonas Uhrig, Nikolaus Mayer, Eddy Ilg,
  Alexey Dosovitskiy, and Thomas Brox.
\newblock Demon: {{Depth}} and motion network for learning monocular stereo.
\newblock In \emph{Proceedings of the {{IEEE}} Conference on Computer Vision
  and Pattern Recognition}, pages 5038--5047, 2017.

\bibitem[Wang et~al.(2021)Wang, Wang, Genova, Srinivasan, Zhou, Barron,
  {Martin-Brualla}, Snavely, and Funkhouser]{wang2021ibrnet}
Qianqian Wang, Zhicheng Wang, Kyle Genova, Pratul~P Srinivasan, Howard Zhou,
  Jonathan~T Barron, Ricardo {Martin-Brualla}, Noah Snavely, and Thomas
  Funkhouser.
\newblock Ibrnet: {{Learning}} multi-view image-based rendering.
\newblock In \emph{Proceedings of the {{IEEE}}/{{CVF}} Conference on Computer
  Vision and Pattern Recognition}, pages 4690--4699, 2021.

\bibitem[Wang et~al.(2019)Wang, Chao, Garg, Hariharan, Campbell, and
  Weinberger]{wang2019pseudo}
Yan Wang, Wei-Lun Chao, Divyansh Garg, Bharath Hariharan, Mark Campbell, and
  Kilian~Q Weinberger.
\newblock Pseudo-lidar from visual depth estimation: Bridging the gap in 3d
  object detection for autonomous driving.
\newblock In \emph{Proceedings of the IEEE/CVF conference on computer vision
  and pattern recognition}, pages 8445--8453, 2019.

\bibitem[Wang et~al.(2023)Wang, Zeng, Guan, Yang, Chen, Liu, Xu, and
  Luo]{wang2023adaptive}
Yuesong Wang, Zhaojie Zeng, Tao Guan, Wei Yang, Zhuo Chen, Wenkai Liu, Luoyuan
  Xu, and Yawei Luo.
\newblock Adaptive patch deformation for textureless-resilient multi-view
  stereo.
\newblock In \emph{Proceedings of the IEEE/CVF Conference on Computer Vision
  and Pattern Recognition}, pages 1621--1630, 2023.

\bibitem[Wei et~al.(2021)Wei, Zhu, Min, Chen, and Wang]{wei2021aa}
Zizhuang Wei, Qingtian Zhu, Chen Min, Yisong Chen, and Guoping Wang.
\newblock Aa-rmvsnet: Adaptive aggregation recurrent multi-view stereo network.
\newblock In \emph{Proceedings of the IEEE/CVF International Conference on
  Computer Vision (ICCV)}, pages 6187--6196, 2021.

\bibitem[Wizadwongsa et~al.(2021)Wizadwongsa, Phongthawee, Yenphraphai, and
  Suwajanakorn]{wizadwongsa2021nex}
Suttisak Wizadwongsa, Pakkapon Phongthawee, Jiraphon Yenphraphai, and Supasorn
  Suwajanakorn.
\newblock Nex: {{Real-time}} view synthesis with neural basis expansion.
\newblock In \emph{Proceedings of the {{IEEE}}/{{CVF}} Conference on Computer
  Vision and Pattern Recognition}, pages 8534--8543, 2021.

\bibitem[Wu et~al.(2014)Wu, Zollh{\"o}fer, Nie{\ss}ner, Stamminger, Izadi, and
  Theobalt]{wu2014real}
Chenglei Wu, Michael Zollh{\"o}fer, Matthias Nie{\ss}ner, Marc Stamminger,
  Shahram Izadi, and Christian Theobalt.
\newblock Real-time shading-based refinement for consumer depth cameras.
\newblock \emph{ACM Transactions on Graphics (ToG)}, 33\penalty0 (6):\penalty0
  1--10, 2014.

\bibitem[Xu and Tao(2019)]{xu2019multi}
Qingshan Xu and Wenbing Tao.
\newblock Multi-scale geometric consistency guided multi-view stereo.
\newblock In \emph{Proceedings of the {{IEEE}}/{{CVF}} Conference on Computer
  Vision and Pattern Recognition}, pages 5483--5492, 2019.

\bibitem[Xu et~al.(2022)Xu, Kong, Tao, and Pollefeys]{xu2022multi}
Qingshan Xu, Weihang Kong, Wenbing Tao, and Marc Pollefeys.
\newblock Multi-scale geometric consistency guided and planar prior assisted
  multi-view stereo.
\newblock \emph{IEEE Transactions on Pattern Analysis and Machine
  Intelligence}, 45\penalty0 (4):\penalty0 4945--4963, 2022.

\bibitem[Yan et~al.(2020)Yan, Wei, Yi, Ding, Zhang, Chen, Wang, and
  Tai]{yan2020dense}
Jianfeng Yan, Zizhuang Wei, Hongwei Yi, Mingyu Ding, Runze Zhang, Yisong Chen,
  Guoping Wang, and Yu-Wing Tai.
\newblock Dense hybrid recurrent multi-view stereo net with dynamic consistency
  checking.
\newblock In \emph{Computer Vision--ECCV 2020: 16th European Conference,
  Glasgow, UK, August 23--28, 2020, Proceedings, Part IV}, pages 674--689.
  Springer, 2020.

\bibitem[Yan et~al.(2018)Yan, Wu, Wang, Xu, An, Guo, and Liu]{yan2018ddrnet}
Shi Yan, Chenglei Wu, Lizhen Wang, Feng Xu, Liang An, Kaiwen Guo, and Yebin
  Liu.
\newblock Ddrnet: {{Depth}} map denoising and refinement for consumer depth
  cameras using cascaded cnns.
\newblock In \emph{Proceedings of the {{European}} Conference on Computer
  Vision ({{ECCV}})}, pages 151--167, 2018.

\bibitem[Yang et~al.(2020)Yang, Mao, Alvarez, and Liu]{yang2020cost}
Jiayu Yang, Wei Mao, Jose~M Alvarez, and Miaomiao Liu.
\newblock Cost volume pyramid based depth inference for multi-view stereo.
\newblock In \emph{Proceedings of the IEEE/CVF Conference on Computer Vision
  and Pattern Recognition}, pages 4877--4886, 2020.

\bibitem[Yang et~al.(2024{\natexlab{a}})Yang, Kang, Huang, Xu, Feng, and
  Zhao]{yang2024depth}
Lihe Yang, Bingyi Kang, Zilong Huang, Xiaogang Xu, Jiashi Feng, and Hengshuang
  Zhao.
\newblock Depth anything: {{Unleashing}} the power of large-scale unlabeled
  data.
\newblock In \emph{Proceedings of the {{IEEE}}/{{CVF}} Conference on Computer
  Vision and Pattern Recognition}, pages 10371--10381, 2024{\natexlab{a}}.

\bibitem[Yang et~al.(2024{\natexlab{b}})Yang, Kang, Huang, Zhao, Xu, Feng, and
  Zhao]{yang2024depth2}
Lihe Yang, Bingyi Kang, Zilong Huang, Zhen Zhao, Xiaogang Xu, Jiashi Feng, and
  Hengshuang Zhao.
\newblock Depth anything {{V2}}.
\newblock \emph{arXiv preprint arXiv:2406.09414}, 2024{\natexlab{b}}.

\bibitem[Yao et~al.(2018)Yao, Luo, Li, Fang, and Quan]{yao2018mvsnet}
Yao Yao, Zixin Luo, Shiwei Li, Tian Fang, and Long Quan.
\newblock Mvsnet: {{Depth}} inference for unstructured multi-view stereo.
\newblock In \emph{Proceedings of the {{European}} Conference on Computer
  Vision ({{ECCV}})}, pages 767--783, 2018.

\bibitem[Yao et~al.(2019)Yao, Luo, Li, Shen, Fang, and Quan]{yao2019recurrent}
Yao Yao, Zixin Luo, Shiwei Li, Tianwei Shen, Tian Fang, and Long Quan.
\newblock Recurrent mvsnet for high-resolution multi-view stereo depth
  inference.
\newblock In \emph{Proceedings of the IEEE/CVF conference on computer vision
  and pattern recognition}, pages 5525--5534, 2019.

\bibitem[Yeshwanth et~al.(2023)Yeshwanth, Liu, Nie{\ss}ner, and
  Dai]{yeshwanth2023scannet++}
Chandan Yeshwanth, Yueh-Cheng Liu, Matthias Nie{\ss}ner, and Angela Dai.
\newblock Scannet++: {{A}} high-fidelity dataset of 3d indoor scenes.
\newblock In \emph{Proceedings of the {{IEEE}}/{{CVF}} International Conference
  on Computer Vision}, pages 12--22, 2023.

\bibitem[Yuan et~al.(2018)Yuan, Gao, Fu, and Xia]{yuan2018temporal}
Ming-Ze Yuan, Lin Gao, Hongbo Fu, and Shihong Xia.
\newblock Temporal upsampling of depth maps using a hybrid camera.
\newblock \emph{IEEE Transactions on Visualization and Computer Graphics},
  25\penalty0 (3):\penalty0 1591--1602, 2018.

\bibitem[Yuan et~al.(2024)Yuan, Cao, Wang, and Li]{yuan2024tsar}
Zhenlong Yuan, Jiakai Cao, Zhaoqi Wang, and Zhaoxin Li.
\newblock Tsar-mvs: Textureless-aware segmentation and correlative refinement
  guided multi-view stereo.
\newblock \emph{Pattern Recognition}, 154:\penalty0 110565, 2024.

\bibitem[Zhang et~al.(2020)Zhang, Ramanagopal, Vasudevan, and
  {Johnson-Roberson}]{zhang2020listereo}
Junming Zhang, Manikandasriram~Srinivasan Ramanagopal, Ram Vasudevan, and
  Matthew {Johnson-Roberson}.
\newblock Listereo: {{Generate}} dense depth maps from lidar and stereo
  imagery.
\newblock In \emph{2020 {{IEEE}} International Conference on Robotics and
  Automation ({{ICRA}})}, pages 7829--7836. IEEE, 2020.

\bibitem[Zhang et~al.(2023{\natexlab{a}})Zhang, Wang, Li, Huang, Sato, and
  Lu]{zhang2023structural}
Mingfang Zhang, Jinglu Wang, Xiao Li, Yifei Huang, Yoichi Sato, and Yan Lu.
\newblock Structural multiplane image: {{Bridging}} neural view synthesis and
  3d reconstruction.
\newblock In \emph{Proceedings of the {{IEEE}}/{{CVF}} Conference on Computer
  Vision and Pattern Recognition}, pages 16707--16716, 2023{\natexlab{a}}.

\bibitem[Zhang et~al.(2023{\natexlab{b}})Zhang, Peng, Hu, and
  Wang]{zhang2023geomvsnet}
Zhe Zhang, Rui Peng, Yuxi Hu, and Ronggang Wang.
\newblock Geomvsnet: {{Learning}} multi-view stereo with geometry perception.
\newblock In \emph{Proceedings of the {{IEEE}}/{{CVF}} Conference on Computer
  Vision and Pattern Recognition}, pages 21508--21518, 2023{\natexlab{b}}.

\bibitem[Zhu et~al.(2021)Zhu, Min, Wei, Chen, and Wang]{zhu2021deep}
Qingtian Zhu, Chen Min, Zizhuang Wei, Yisong Chen, and Guoping Wang.
\newblock Deep learning for multi-view stereo via plane sweep: {{A}} survey.
\newblock \emph{arXiv preprint arXiv:2106.15328}, 2021.

\end{thebibliography}
}
\clearpage
\appendix

\section{Details on Losses \& Regularizers}\label{suppl:loss_reg}

In this section, we describe the geometric, photometric and dense geometric consistency losses and regularizers that are used to control and stabilize our optimization in more detail.

\paragraph*{Geometric Loss.}
$\lossgeo$ measures the distance between the current depth buffer and the depth values of the projected sparse point cloud.
It is defined as
\begin{equation}
\small 
 \lossgeo(\D_0^*, \D_0^c,\X) = l_\textit{geo}(\D_0^*,\X) + l_\textit{geo}(\D_0^c,\X), \textrm{with} 
\end{equation}
\begin{equation}
\small 
    l_\textit{geo}(\D,\X) = \frac{1}{|\X|} \sum_{j=1}^{|\X|} \frac{\mathcal{H}_\delta(z^\D_j, z^\X_j)}{z^\X_j}.
\end{equation}
$\mathcal{H}$ is the Huber loss~\cite{huber1992robust} with $\delta = 0.5$.
$|\X|$ is the number of points available in $\X$.
$z^\X_j$ are the available depth values from the sparse point cloud $\X$, projected to the current view, and $z^\D_j$ the corresponding depth values of $\D_0^*$ or $\D_0^c$, respectively, evaluated at the image space coordinates of $\X$.
$l_\textit{geo}(\D_0^c,\X)$ will only come into effect during the coarse refinement.

\paragraph*{Photometric Loss.}
We use a standard mean squared error formulation for our photometric consistency loss $\lossmse$ between rendered $\I^*_i$ and captured images $\I_i$:
\begin{equation}
\small 
    \lossmse(\I^*_i,\I_i) = ||\max(0,\E_i^*)(\I^*_i - \I_i)||_2.
\end{equation}
$\max(0,\E_i^*)$ indicates the projected edge map $\E_r^*$ will mask out regions of $\I_i$ which were occluded in $\I_0$.

\paragraph*{Regularizers.}

For training regularization, we apply three different regularizers.

The regularizers $\regpoisson$ and $\regedge$ have the goal to maintain structural information from the mono depth image $\D_0^r$.
To this end, we calculate edge images~$\E_0^r$ and $\E_0^*$ from $\D_0^r$ and $\D_0^*$ by utilizing a Sobel filter-based edge detection.
The edge images $\E$ have values $\in [-1,1]$, with values below  0 indicating edges and discontinuities and values above 0 indicating surfaces.

Following, we define the edge regularizer $\regedge$ as 
\begin{equation}
\small 
\begin{array}{c}
    \regedge(\E_0^*, \E_0^r) = ||\max(0,\E_0^*) - \max(0,\E_0^r)||_2
    + \frac{1}{|\M|} \sum\limits_{j=1}^{|\M|} \mathbf{e}_{u',v'}
\end{array}
\end{equation}
where $\mathbf{e}_{u',v'}$ are the sampled values of $\E_0^r$ at image coordinates $u',v'$ and $|\M|$ is the amount of vertices present in the mesh $\M$.
The first term of $\regedge$ encourages that silhouettes of $\D_0^*$ and $\D_0^r$ align without enforcing the absolute values to match (thus, clamping the edge images at 0).
The second term encourages the offsets~$\Delta u,\Delta v$ to move the vertices' image coordinates to the midpoint of the edges.

Inspired by Poisson blending~\cite{perez2003poisson}, we define $\regpoisson$ as:
\begin{equation}
\small 
\begin{array}{c}
     \regpoisson(\D_0^*,\D_0^r) = ||\max(0,\E_0^r)(\nabla_u \Tilde{\D}_0^* - \nabla_u \D_0^r)||_2 + \\\\
     ||\max(0,\E_0^r)(\nabla_v \Tilde{\D}_0^* - \nabla_v \D_0^r)||_2.
\end{array}
\end{equation}
$\Tilde{\D}_0^*$ is $\D_0^*$ normalized to [0,1].
$\nabla_u$ and $\nabla_v$ are the image space gradient operators in $u$ and $v$ direction.
For $\regpoisson$, we use $\E_0^r$ to mask out silhouettes as we only want to transfer local structural details while preserving absolute scale while neglecting bigger discontinuities, which are prone to be inaccurate.

Our total evaluation loss combines the introduced objective functions using a weighted sum of each of the elements, see Tab.~\ref{tab:loss_weigths}.

\begin{table}[h]
\centering
\small
\begin{tabular}{lrl|lr}
\multicolumn{2}{l}{Loss} &&  \multicolumn{2}{l}{Regularizer} \\
\hline
$\lossmse$     & 1.0     && $\regpoisson$      & 400.0      \\
$\lossgeo$     & 0.1     && $\regedge$         & 0.1        \\
($\lossmse^\text{MV}$     & 1.0)        && ($\regsmooth$       & 0.01)    \\
($\lossgeomconsistency$     & 0.1)        && ($\mathcal{R}_n$       & 0.1)   
\end{tabular}
\caption{Used weights for computing the total optimization loss. Brackets indicate usage only in ablation studies. }
    \label{tab:loss_weigths}
\end{table}

\section{Training Details}
\label{suppl:impl_details}

For our implementation, we use the hardware accelerated framework NVDiffRast~\cite{Laine2020diffrast} with the adjusted formulation outlined in Sec.~\ref{suppl:nvdiffrast_support}. 
For initial depth estimations, we use Marigold~\cite{ke2024repurposing}, if not stated otherwise, and employ COLMAP~\cite{schonberger2016structure,schoenberger2016mvs} for Structure-from-Motion and image undistortion with configurations as recommended by the used dataset~\cite{yeshwanth2023scannet++}.
In Fig.~\ref{fig:pcds}, we depict sparse pointclouds of representative views, showcasing sparseness especially on white walls.
For coarse refinement, $\Psi$ is a shallow MLP with degrees of positional encoding being $m=3$ and $k=5$. Two hidden layers of width 16, initialized with a normal distribution with $\sigma=0.1$. 
This comparatively small initialization ensures $o_g$ and $s_g$ being very small initially, thus, starting close to the initial depth estimation.
During the coarse refinement phase, we used a learning rate of $0.001$ and stop after 400 iterations with batch sizes of 16. 
The current states of $o_g$ and $s_g$ are baked into $\M$'s $z$.
We stop optimization after another 1100 iterations of local refinement with learning rates of $0.0005$. 
The number of iterations were found empirically based on our ScanNet++ validation set.
In total, the process takes around 8\,min per depth map (including output writing and logging) on a NVIDIA RTX 3090 using the CUDA backend.
We see potential for performance improvements by e.g. replacing the MLP of the coarse refinement by an explicit data structure that can be optimized faster.
We used the Adam~\cite{kingma2014adam} optimizer with default parameters as proposed by the PyTorch framework~\cite{stevens2020deep}.

\section{Implementation with GPU Acceleration}
\label{suppl:nvdiffrast_support}

\newcommand{\far}{z_\textit{far}}
\newcommand{\near}{z_\textit{near}}

Using a GPU hardware accelerated framework (we use NVDiffRast~\cite{Laine2020diffrast}) is preferable due to implementation simplicity and optimization speed.
To leverage hardware acceleration, our implementation must follow specifications as defined by the OpenGL convention~\cite{buss20033d}.
As a consequence, we must fulfill the following conditions:
1) The intrinsics matrix $C$ must be reformulated as a $4\times4$ projection matrix~\cite{buss20033d}, which limits the visible depth range by to the range [$\near, \far$] and normalizes image space positions to $[-1,1]^3$, referred to as \textit{clip space}~\cite{buss20033d}.
2) Consequently, we also need to take care of edge case handling when depth values approach these near and far values.
3) As the coordinate refinement of $(u,v)$ also has to be performed in clip space, we additionally need to store the depth of $\M$ in non-linear reciprocal space. 
Optimizing in reciprocal space would be very instable, so we perform depth refinement distinctly using a linear scaling, as described in Sec.~\ref{sec:local_refinement}.

In the following, we describe the adaptions made to comply with the mentioned requirement in more detail.

\paragraph*{Projection Matrices.}

We convert the intrinsics definition $C$ with $\{f_x,f_y,c_x,c_y\}$ to the projection matrix $P$ following the OpenGL convention~\cite{buss20033d}, with $\near=0.1$ and $\far=100$ using homogeneous 4D coordinates for all vertices by
\begin{equation}
 P = 
 \begin{small}
     \begin{pmatrix}
     \frac{2f_x}{W} & 0 & 1 - \frac{2c_x}{W} & 0 \\
     0 & -\frac{2f_y}{H} & 1- \frac{2c_y}{H}  & 0 \\
     0 & 0 & \frac{-(\far + \near)}{\near - \far} & -\frac{2 \far \near}{\far-\near} \\
     0 & 0 & -1 & 0
     \end{pmatrix}.
 \end{small}
 \end{equation}
 
This $P_{\{0..N\}}$ is then used in place of $C_{\{0..N\}}$ in Eq.~\ref{eq:rendering_projection}.

Adjusting to this projection scheme and its reciprocal depth formulation for depth refinement however requires several adjustments to our method:

\paragraph*{Meshing.}
The $z$ coordinates of $\M$ are stored in reciprocal space, given by
\begin{equation}
\small
    z' = \frac{\far + \near}{\near - \far} +  \frac{2 \far \near}{-z^g(\near - \far)}.
\end{equation}
$u$ and $v$ are stored in normalized image coordinates in the $[-1,1]$ range with 
\begin{equation}
\small
    (u',v') = \left(\frac{2u^g}{W}-1, \frac{2v^g}{H}-1 \right) 
\end{equation}
An upside of this normalization is that it improves numerical stability and also simplifies positional MLP encoding (see Sec.~\ref{sec:coarse_refinement}).

\paragraph*{Depth and Vertex Refinement.}
For both global and local refinement, depth optimization in reciprocal space leads to instable optimization.
As such we do this optimization in linear space and substitute the rightmost part in Eq.~\ref{eq:rendering_projection} first by
\begin{equation}
\small 
        p'' = \frac{1+s_c}{f_z}P_0^{-1}((u'+\Delta u,v'+\Delta v,z',1)^T +o_c.
\end{equation}

Regarding the depth, we also need to prevent vertices from moving outside the view frustum by
\begin{equation}
\small 
\begin{array}{c}
   p' = \min(-\frac{\far-\epsilon}{z}, 1) \max(-\frac{\near+\epsilon}{z}, 1) p''.
\end{array}
\end{equation}
with $\epsilon$ empirically set to $10^{-6}$.
As such we do depth optimization in linear space while keeping image coordinate optimization in projected space.

Thus, Eq.~\ref{eq:rendering_projection} can be rewritten as: 
\begin{equation}
\small 
\label{eq:opengl_optimization}
    p^*_i = P_i [R_i|t_i]  [R_0|t_0]^{-1} p',
\end{equation}
with $[R|t]$ matrices now being $4\times4$ dimensional, respectively.

\paragraph*{Tonemapping}
\label{suppl:tonemapping}

Camera color ranges often vary with changing lighting, affecting the reliability of photometric losses. To simulate these variations, we use a method akin to analysis-by-synthesis by modeling the camera's response curve and exposure settings. 
Consequently, we incorporated a differential tonemapping module, inspired by Rückert et al.~\cite{ruckert2022adop} and common to NVS~\cite{hahlbohm2024inpc,franke2024trips}.

In contrast to their implementation, our tonemapping function needs to be inverted.
Thus, we opted for a trivially invertible gamma-curve instead of using a piecewise linear function to model the camera response curve.

\section{Dataset preparation} 
\label{suppl:datasets}

\paragraph*{ScanNet++.}

\begin{figure*}
    \centering
    \includegraphics[height=0.17\linewidth]{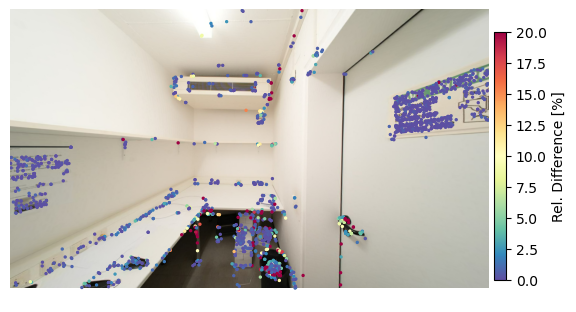}
    \includegraphics[height=0.17\linewidth]{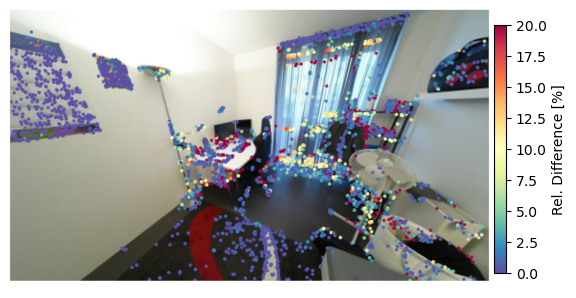}
    \includegraphics[height=0.17\linewidth]{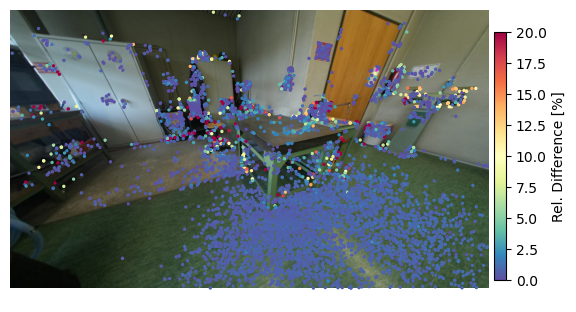}
    \caption{Visualization of the \textit{visible} sparse pointclouds from SfM of diffferent representative views. Color of the points encodes the relative depth error compared to the groundtruth depth map. Number of points and standard deviation of the relative error was, from left to right: 1.7k and 9.2\,\%, 3.3k and 5.7\,\%, and 6k and 3.5\,\%.}
    \label{fig:pcds}
\end{figure*}

For the ScanNet++ dataset, ground truth depth maps are rendered from the meshes, for the sparse point cloud, poses, and undistortion, we employed COLMAP~\cite{schonberger2016structure} as proposed by the  dataset toolkit~\cite{yeshwanth2023scannet++}.

For the results present in Main-Sec.~\ref{sec:results_related}, we used 10 scenes from the ScanNet++ validation set and 10 depth maps of each (100 in total).
These scenes were chosen to have 300 consecutive frames without cuts, as we chose auxiliary views based on the temporal neighborhood. Additionally, we only considered scenes with less than 15\,\% of the views being marked as blurry.
To tune the hyperparameters of our approach~(see Sec.~\ref{suppl:ablation_studies}), such as MLP configuration or decimation, we prepared another, different seven scenes with 10 views each (70 in total) in an analogous manner.
Note that these seven scenes have not been used in the evaluations of the main paper.
The hyperparameters found in the ablation studies were used consistently for all results showcased in the Main-Sec.~\ref{sec:results_related}.

\paragraph*{Replica.}

For the Replica dataset, color and ground truth depth maps are obtained from the rendered meshes.
We selected four scenes, and optimized over 64 views for each from the subset prepared by Rosinol et al.~\cite{rosinol2023nerf}.
As sparse point clouds, we generated 5000 point samples from the ground truth depth map, added a relative error with standard deviation of 2\%, and randomly replaced 2\% by randomly distributed outliers from the overall depth range.
To examine the impact of the quality of the point cloud, we also generated worse point clouds: for two scenes (\texttt{room0} and \texttt{room2}) we went down to 3500 points with relative error of 5\% and 5\% outliers, and for scenes \texttt{office1} and \texttt{office4} only 1000 points were generated, with a relative error of 10\% and 10\% outliers.
The numbers were chosen to approximately match the error distribution as observed in real-world scenarios, see Fig.~\ref{fig:pcds}.
Nevertheless, the uniform density of the constructed sparse point clouds is unrealistic which may limit expressiveness.

\paragraph*{Number of views.}
To optimize for photo-consistency, we used 60 auxiliary views for the Replica dataset and 30 for ScanNet++. 
Auxiliary views were simply the set of consecutive frames which are temporally closest to $\I_0$. 
MVSFormer, MVSFormer++ and GeoMVSNet received their default amount of 10 views as input which were evenly distributed across the set of our auxiliary views.

Note that our approach works best with a high number of auxiliary views with a distribution that can be considered as a front facing capture, while input views with extreme baselines offer little benefits due to the masking of regions with potential disocclusions. 
A higher number of auxiliary views will generally increase robustness and accuracy of the optimization.
The number of used input views was thus empirically set according to the estimated framerate of the provided dataset, covering approximately a 1-2\,s video clip. (Replica has approximately 30 frames per second, ScanNet++ ~10 - 15, exact timestamps are, to the best of our knowledge, not provided).

\paragraph*{Tanks and Temples.}
Regarding the Tanks and Temples results, we used every 5th frame of the original video for Ignatius, every 15th frame for Courthouse and every 30th frame for Truck. 
The views for our multi-view optimization were simply selected by using the 30, 15, and 7 preceding and succeeding frames (for Ignatius, Courthouse, and Truck respectively).
We used the same view selection for MVSFormers. 
For point cloud fusion, we stuck to the implementation by Cao et al.~\cite{Cao2022MVSFormerMS} (MVSFormer).
This implementation receives a confidence map and a respective threshold as input, which we set to 0.8.
To fuse our pointclouds, we used the edge map $\mathbf{E}$ as confidence map.

\section{Additional Results}
\label{suppl:add_results}

\paragraph*{Qualitative results on Replica.}

\begin{figure}
    \centering
    \includegraphics[width=0.99\linewidth]{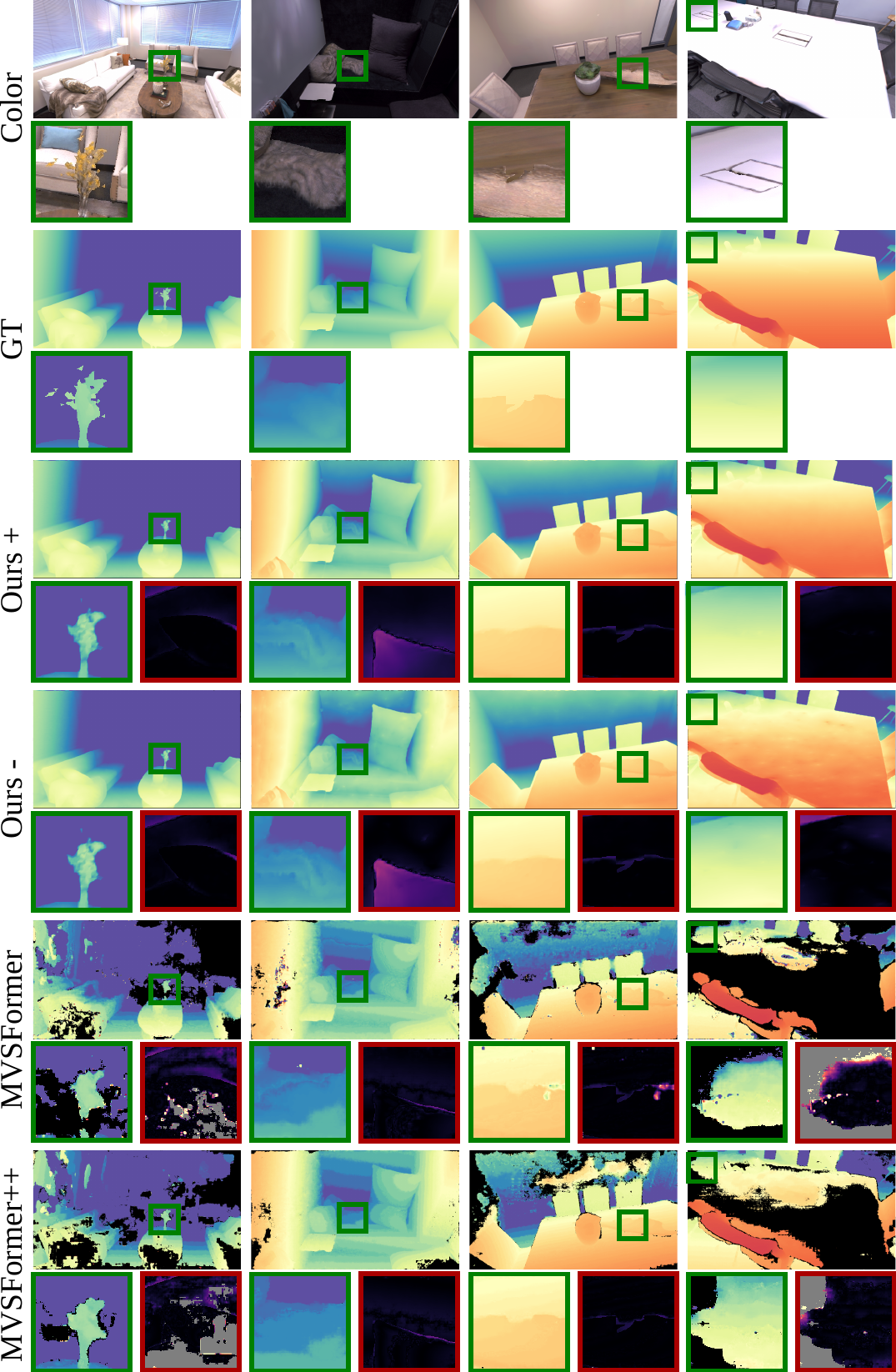}\vspace{-2mm}
    \caption{Results on the Replica dataset with MVSFormers' certainty threshold at their default of 0.5. Green borders indicate image crops, while red highlights error maps; brighter areas imply greater errors, and grey indicates missing samples.}
    \label{fig:eval_replica}
\end{figure}

Visual examples of Replica results are shown in Fig.~\ref{fig:eval_replica}. 
Noise like artifacts (see e.g the pillow case and wall in the center column) can emerge when there are severe outliers in the sparse point cloud as in the case of 'Ours~(-)'.
Generally, our method produces dense and clean results also for feature scarce areas, where others struggle. 

\paragraph*{Full scene reconstruction on Tanks and Temples.}

\begin{table}[]
\tiny
\addtolength{\tabcolsep}{-3pt}
\begin{tabular}{l|ccc|ccc|ccc}
 Method & 
  \multicolumn{3}{c|}{Ignatius \quad (t= 0.003)} &
  \multicolumn{3}{c|}{Courthouse \quad (t= 0.025)} &
  \multicolumn{3}{c}{Truck  \quad (t= 0.005)} \\
 \quad \quad Metric $\uparrow$ &
  {(P)recision} &
  {(R)ecall} &
  {(F)-Score} &
  {P} &
  {R} &
  {F} &
  P &
  R &
  F \\\hline
MVSFormer &
  \cellcolor[HTML]{C9E9D9}0.59 &
  \cellcolor[HTML]{57BB8A}0.97 &
  \cellcolor[HTML]{79C9A2}0.74 &
  \cellcolor[HTML]{6AC397}0.48 &
  \cellcolor[HTML]{57BB8A}0.60 &
  \cellcolor[HTML]{62C092}0.53 &
  \cellcolor[HTML]{57BB8A}0.74 &
  \cellcolor[HTML]{DCF1E6}0.59 &
  \cellcolor[HTML]{91D3B3}0.66 \\
MVSFormer++ &
  \cellcolor[HTML]{ABDDC4}0.62 &
  \cellcolor[HTML]{5DBE8F}0.96 &
  \cellcolor[HTML]{57BB8A}0.75 &
  \cellcolor[HTML]{57BB8A}0.51 &
  \cellcolor[HTML]{5FBF90}0.59 &
  \cellcolor[HTML]{57BB8A}0.55 &
  \cellcolor[HTML]{7BCAA3}0.68 &
  \cellcolor[HTML]{57BB8A}0.78 &
  \cellcolor[HTML]{57BB8A}0.73 \\
COLMAP &
  \cellcolor[HTML]{57BB8A}0.69 &
  \cellcolor[HTML]{FFFFFF}0.72 &
  \cellcolor[HTML]{B9E3CE}0.71 &
  \cellcolor[HTML]{FFFFFF}0.25 &
  \cellcolor[HTML]{FFFFFF}0.47 &
  \cellcolor[HTML]{FFFFFF}0.33 &
  \cellcolor[HTML]{A1D9BD}0.62 &
  \cellcolor[HTML]{B9E3CF}0.64 &
  \cellcolor[HTML]{A5DBC0}0.63 \\
Ours &
  \cellcolor[HTML]{FFFFFF}0.55 &
  \cellcolor[HTML]{A2DABE}0.86 &
  \cellcolor[HTML]{FFFFFF}0.67 &
  \cellcolor[HTML]{A9DDC3}0.38 &
  \cellcolor[HTML]{F2FAF6}0.48 &
  \cellcolor[HTML]{B5E1CB}0.43 &
  \cellcolor[HTML]{EEF9F4}0.50 &
  \cellcolor[HTML]{FFFFFF}0.54 &
  \cellcolor[HTML]{FEFFFE}0.52 \\
Ours$_\text{DA}$ &
  \cellcolor[HTML]{BAE3CF}0.61 &
  \cellcolor[HTML]{B4E1CB}0.83 &
  \cellcolor[HTML]{BFE5D3}0.70 &
  \cellcolor[HTML]{B1E0C9}0.37 &
  \cellcolor[HTML]{FEFFFE}0.47 &
  \cellcolor[HTML]{BDE5D1}0.41 &
  \cellcolor[HTML]{FFFFFF}0.47 &
  \cellcolor[HTML]{EBF7F1}0.57 &
  \cellcolor[HTML]{FFFFFF}0.52
\end{tabular}\vspace{-2mm}
\caption{\label{tab:eval_tnt}Results on the Tanks and Temples~\cite{knapitschTanksTemplesBenchmarking2017a} dataset. Ours$_\text{DA}$ indicates usage of DepthAnything2~\cite{yang2024depth} instead of Marigold~\cite{ke2024repurposing}.}
\end{table}

\begin{figure}
    \centering
    \includegraphics[width=\linewidth]{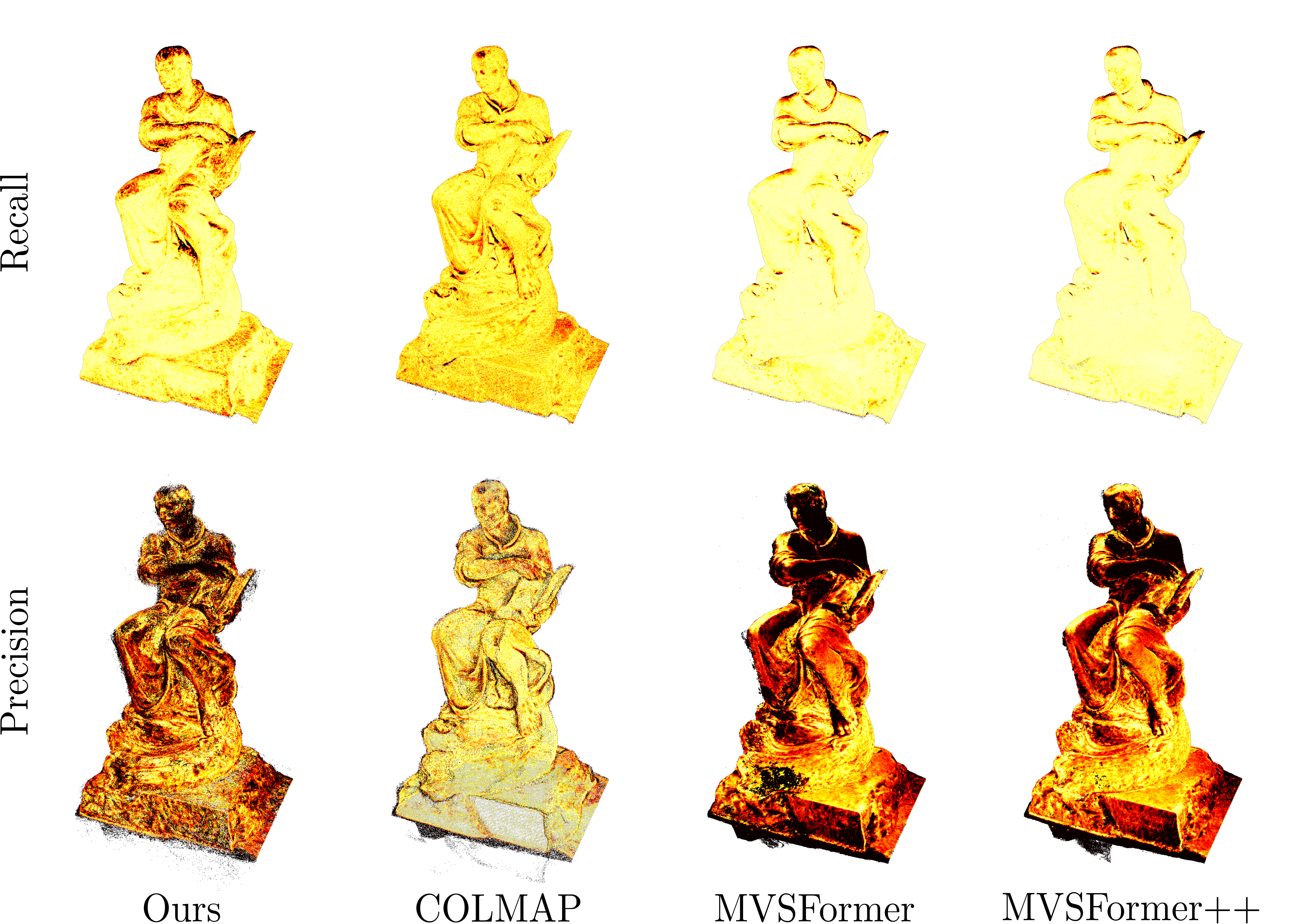}
    \caption{Qualitative results of the Tanks and Temples Scene Ignatius. Brighter is better}
    \label{fig:ignatz}
\end{figure}

Additionally, to individual depth map comparisons, we evaluated on the task of full scene reconstruction which implies a fusion and filtering process of involved depth maps. 
In Tab.~\ref{tab:eval_tnt}, we present the scores of the dense point cloud reconstructions following the standard evaluation protocol of the Tanks and Temples reconstruction benchmark~\cite{knapitschTanksTemplesBenchmarking2017a} and 
visual results are presented in Fig.~\ref{fig:ignatz}.

We used COLMAP's default settings for the dense reconstruction used the resulting SfM poses for all other methods. 
For generating both MVSFormers' and our results, we employed Yao et al.'s point cloud fusion method~\cite{yao2018mvsnet}.

Again we used the same configuration and hyperparameters as in the main paper to produce our results, with two exceptions: We changed the learning rate (1e-3 $\rightarrow$ 5e-4) during coarse alignment of the Truck scene.
In this case, the scene scale was very different compared to the other scenes.
Secondly, we applied the tonemapper presented in the main paper to account for challenging exposure changes. 
An ablation study is presented in Suppl.-Sec.~\ref{suppl:ablation_studies}.
We incorporate results using both Marigold and DepthAnything2 as input to our method, since we found that Marigold usually performs better indoors, while DepthAnything2 is favorable for outdoor settings.

We observe that our method performs worse on the Tanks and Temples dataset compared to the state-of-the-art. 
In contrast to MVSFormer and MVSFormer++, which are mainly trained on semantically related scenes, our method was not adapted for or tuned to the Tanks and Temples dataset and outperforms the same methods on the other datasets. 
In addition, Tanks and Temples only provides point clouds as ground truth and therefore many advantages of our method, like high completeness per depth map, are not beneficial in this evaluation.

\section{Ablation Studies}
\label{suppl:ablation_studies}

In this section, we evaluate different choices of hyper parameter settings and present ablation studies based on seven withheld scenes of the ScanNet++ dataset, referred to as the ablation or withheld scene set.

\paragraph*{Loss Ablations.}

\begin{table}[]
\centering
\tiny
\addtolength{\tabcolsep}{-2pt}
\begin{tabular}{l|rrrr|rrr}
Configuration &
  \multicolumn{1}{l}{RMSE} &
  \multicolumn{1}{l}{MAE} &
  \multicolumn{1}{l}{L1-rel} &
  \multicolumn{1}{l|}{L1-inv} &
  \multicolumn{1}{c}{$\uparrow$Acc.} &
  \multicolumn{1}{c}{Acc.} &
  \multicolumn{1}{c}{Acc.} \\
    & \multicolumn{1}{c}{$\downarrow$}
    &
    &
    &
    &
  \multicolumn{1}{c}{0.01\,m} &
  \multicolumn{1}{c}{0.05\,m} &
  \multicolumn{1}{c}{0.10\,m} \\\hline
$\scriptstyle\lossgeo$ &
  \cellcolor[HTML]{E67C73}0.32 &
  \cellcolor[HTML]{E78077}0.15 &
  \cellcolor[HTML]{E67C73}0.27 &
  \cellcolor[HTML]{EB928B}0.13 &
  \cellcolor[HTML]{E67C73}0.14 &
  \cellcolor[HTML]{EFABA5}0.47 &
  \cellcolor[HTML]{F4C5C1}0.69 \\
$\scriptstyle\lossgeo + \lossmse $ &
  \cellcolor[HTML]{EA9189}0.31 &
  \cellcolor[HTML]{E67C73}0.15 &
  \cellcolor[HTML]{FBE7E6}0.14 &
  \cellcolor[HTML]{E67C73}0.14 &
  \cellcolor[HTML]{EFADA8}0.16 &
  \cellcolor[HTML]{E67C73}0.42 &
  \cellcolor[HTML]{E67C73}0.59 \\
$\scriptstyle\lossgeo+ \regpoisson$ &
  \cellcolor[HTML]{F9DCD9}0.24 &
  \cellcolor[HTML]{FBE8E7}0.11 &
  \cellcolor[HTML]{FEF6F5}0.12 &
  \cellcolor[HTML]{F8D8D5}0.11 &
  \cellcolor[HTML]{FFFFFF}0.18 &
  \cellcolor[HTML]{FCF3F2}0.55 &
  \cellcolor[HTML]{FDF8F7}0.76 \\
$\scriptstyle\lossgeo+  \regpoisson + \regedge$ &
  \cellcolor[HTML]{FFFFFF}0.21 &
  \cellcolor[HTML]{FFFFFF}0.10 &
  \cellcolor[HTML]{FFFFFF}0.11 &
  \cellcolor[HTML]{FFFFFF}0.10 &
  \cellcolor[HTML]{FDF5F4}0.17 &
  \cellcolor[HTML]{FFFFFF}0.57 &
  \cellcolor[HTML]{FFFFFF}0.76 \\
$\scriptstyle\lossgeo + \mathcal{L}_\textit{photo,(c\&l)} + \regpoisson + \regedge$ &
  \cellcolor[HTML]{57BB8A}0.18 &
  \cellcolor[HTML]{57BB8A}0.08 &
  \cellcolor[HTML]{57BB8A}0.08 &
  \cellcolor[HTML]{5CBD8E}0.09 &
  \cellcolor[HTML]{57BB8A}0.27 &
  \cellcolor[HTML]{57BB8A}0.64 &
  \cellcolor[HTML]{61BF91}0.80 \\
\color{blue}$\scriptstyle\lossgeo + \mathcal{L}_\textit{photo} + \regpoisson + \regedge$ &
  \cellcolor[HTML]{A0D8BD}0.19 &
  \cellcolor[HTML]{78C8A0}0.09 &
  \cellcolor[HTML]{82CCA8}0.08 &
  \cellcolor[HTML]{57BB8A}0.09 &
  \cellcolor[HTML]{5FBF90}0.27 &
  \cellcolor[HTML]{65C194}0.64 &
  \cellcolor[HTML]{57BB8A}0.80 
\end{tabular}\vspace{-2mm}
    \caption{Ablation study on used training losses, with $\mathcal{L}_\textit{photo,(c\&l)}$ indicates applying photometric loss also in coarse refinement instead of only local refinement.
    Our default configuration is in {\color{blue}blue}.
    }
    \label{tab:ablation_losses}
\end{table}

\begin{figure}
    \centering
    \includegraphics[width=0.98\linewidth]{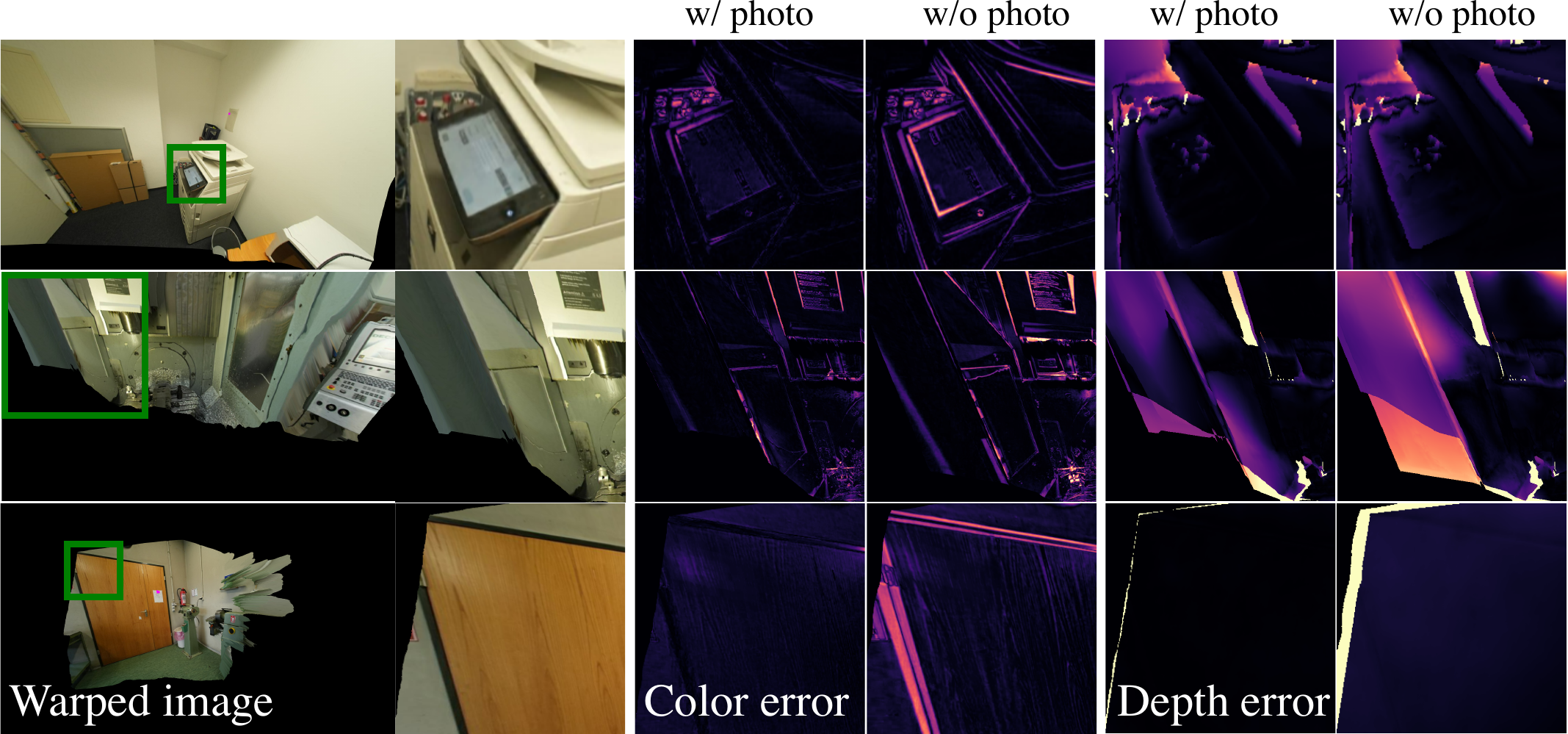}\vspace{-1mm}
    \caption{Results of depth maps under warping. 'w/ photo' indicates use of $\lossmse$, 'w/o photo' omits this. Error maps shown on the right; brighter areas imply greater errors.}
    \label{fig:warperror}
\end{figure}

As presented in Tab.~\ref{tab:ablation_losses}, using only $\lossgeo$~(see Suppl.~Sec.~\ref{suppl:loss_reg}) performs subpar.
When $\lossgeo$ gets combined with the proposed regularizers, distance metrics like RMSE, MAE, or L1-rel are improved. 
Adding $\lossmse$ during local refinement, notably boosts accuracy measures (30\,\% relative improvement in case of $\tau$=0.01\,m and 11\,\% in case of 0.05\,m).
Furthermore, as seen in Fig.~\ref{fig:warperror}, photometric loss is able to finely adjust inaccuracies in the depth map.

While aiming for photometric consistency also during coarse refinement (indicated by $\mathcal{L}_\textit{photo,(c\&l)}$) increases the risk of getting stuck in local minima, this proved irrelevant in the case of the ablation scene set.

\paragraph*{Photometric Loss Choice.}
Ablation on the withheld scene set, extends Tab.~\ref{tab:ablation_losses}.
while invariances of the VGG loss did not guide towards precise local matching, adding a perceptual loss like SSIM performed on par or improved metrics, underlining the effectiveness of photometric consistency as an objective function.

\begin{table}[H]
{
\centering
\tiny
\addtolength{\tabcolsep}{-2.5pt}
\centering
\begin{tabular}{l|rrrrrrr}
&
  \multicolumn{1}{l}{RMSE} &
  \multicolumn{1}{l}{MAE} &
  \multicolumn{1}{l}{L1-rel} &
  \multicolumn{1}{l|}{L1-inv} &
  \multicolumn{1}{c}{$\uparrow$Acc.} &
  \multicolumn{1}{c}{Acc.} &
  \multicolumn{1}{c}{Acc.} \\
    & \multicolumn{1}{c}{$\downarrow$}
    &
    &
    &
    &
  \multicolumn{1}{c}{0.01\,m} &
  \multicolumn{1}{c}{0.05\,m} &
  \multicolumn{1}{c}{0.10\,m} \\\hline
{\color{blue}Default losses} &
  \textbf{0.18} &
  \textbf{0.08} &
  \textbf{0.08} &
  \textbf{0.09} &
  0.27 &
  0.64 &
  0.80 \\
+SSIM & \textbf{0.18} & \textbf{0.08} & 0.10 & 0.10 & \textbf{0.36} & \textbf{0.69} & \textbf{0.82} \\
+VGG  & 0.25          & 0.13          & 0.14 & 0.15 & 0.18          & 0.48          & 0.65         
\end{tabular}
}
\caption{
    Configuration used to produce results of the main paper is marked in {\color{blue}blue}.}
\end{table}

\paragraph*{Alternative Regularizers.}


\begin{table}[]
\centering
\tiny
\addtolength{\tabcolsep}{-2pt}
\begin{tabular}{l|rrrr|rrr}
Configuration &
  \multicolumn{1}{l}{RMSE} &
  \multicolumn{1}{l}{MAE} &
  \multicolumn{1}{l}{L1-rel} &
  \multicolumn{1}{l|}{L1-inv} &
  \multicolumn{1}{c}{Acc.$\uparrow$} &
  \multicolumn{1}{c}{Acc.} &
  \multicolumn{1}{c}{Acc.} \\
   & \multicolumn{1}{c}{$\downarrow$}
  &
  &
  &
 &
  \multicolumn{1}{c}{0.01\,m} &
  \multicolumn{1}{c}{0.05\,m} &
  \multicolumn{1}{c}{0.10\,m} \\\hline
$\scriptstyle\lossgeo + \lossmse + \regsmooth$ &
  \cellcolor[HTML]{E67C73}0.391 &
  \cellcolor[HTML]{E67C73}0.194 &
  \cellcolor[HTML]{E67C73}0.296 &
  \cellcolor[HTML]{E67C73}0.161 &
  \cellcolor[HTML]{E67C73}0.152 &
  \cellcolor[HTML]{E67C73}0.414 &
  \cellcolor[HTML]{E67C73}0.582 \\
$\scriptstyle\lossgeo + \lossmse + \mathcal{R}_\textit{n} $ &
  \cellcolor[HTML]{FFFFFF}0.188 &
  \cellcolor[HTML]{57BB8A}0.080 &
  \cellcolor[HTML]{FFFFFF}0.080 &
  \cellcolor[HTML]{FFFFFF}0.088 &
  \cellcolor[HTML]{FFFFFF}0.193 &
  \cellcolor[HTML]{FFFFFF}0.589 &
  \cellcolor[HTML]{57BB8A}0.805 \\
\color{blue}$\scriptstyle\lossgeo + \mathcal{L}_\textit{photo} + \regpoisson + \regedge$ &
  \cellcolor[HTML]{57BB8A}0.177 &
  \cellcolor[HTML]{FFFFFF}0.082 &
  \cellcolor[HTML]{57BB8A}0.077 &
  \cellcolor[HTML]{57BB8A}0.088 &
  \cellcolor[HTML]{57BB8A}0.273 &
  \cellcolor[HTML]{57BB8A}0.644 &
  \cellcolor[HTML]{FFFFFF}0.801
\end{tabular}
    \caption{Ablation study on used smoothing or gradient-based regularizers. 
    Ours is marked in {\color{blue}blue}.
    }
    \label{tab:ablation_regularizer}
\end{table}

Tab.~\ref{tab:ablation_regularizer} shows results for replacing our Poisson regularizer $\regpoisson$ and edge-aware regularizer $\regedge$ by a very simple smoothness term $\regsmooth$ or using an estimated normal map~\cite{bae2024rethinking} for regularization indicated by $\mathcal{R}_\textit{n}$.

The $\regsmooth$ is a simple regularizer that suppresses noisy results:
\begin{equation}
\small 
    \regsmooth(\D_0^*) = ||\D_0^* - g(\D_0^*)||_2,
\end{equation}
where $g$ is a Gaussian blur with window size 5.

$\mathcal{R}_\textit{n}$ measures the cosine similarity of the Sobel-based normal of $\D^*_0$ with the estimate of the normal predictor~\cite{bae2024rethinking}.

Our choice of regularizers significantly outperforms $\regsmooth$ demonstrating the importance of more sophisticated regularization.
Simultaneously, our regularizers are on par with $\mathcal{R}_\textit{n}$.

\paragraph*{Coarse \& Local Refinement}

\begin{table}[]
\centering
\tiny
\addtolength{\tabcolsep}{-2.5pt}
\begin{tabular}{l|rrr|rrrr}
Setup &
  \multicolumn{1}{c}{RMSE} &
  \multicolumn{1}{c}{MAE} &
  \multicolumn{1}{c}{L1-rel} &
  \multicolumn{1}{c}{$\delta_1$} &
  \multicolumn{1}{c}{Acc.} &
  \multicolumn{1}{c}{Acc.} &
  \multicolumn{1}{c}{Acc.} \\
    &\multicolumn{1}{c}{$\downarrow$}
    &
    &
    &\multicolumn{1}{c}{$\uparrow$}
    &
  \multicolumn{1}{c}{0.01\,m} &
  \multicolumn{1}{c}{0.05\,m} &
  \multicolumn{1}{c}{0.10\,m} \\\hline
Coarse: none; Local: default &
  \cellcolor[HTML]{FCF0EE}0.23 &
  \cellcolor[HTML]{FAE3E2}0.14 &
  \cellcolor[HTML]{FCEDEB}0.15 &
  \cellcolor[HTML]{F6D0CD}0.82 &
  \cellcolor[HTML]{F1BAB5}0.14 &
  \cellcolor[HTML]{F1B5B0}0.39 &
  \cellcolor[HTML]{F3C0BC}0.58 \\
Coarse: $(o^g,s^g)$; Local: default &
  \cellcolor[HTML]{57BB8A}0.16 &
  \cellcolor[HTML]{57BB8A}0.08 &
  \cellcolor[HTML]{57BB8A}0.08 &
  \cellcolor[HTML]{FFFFFF}0.91 &
  \cellcolor[HTML]{FFFFFF}0.24 &
  \cellcolor[HTML]{FFFFFF}0.61 &
  \cellcolor[HTML]{FFFFFF}0.77 \\
{\color{blue}Coarse: MLP$_\text{S}$; Local: default} &
  \cellcolor[HTML]{B9E2CE}0.18 &
  \cellcolor[HTML]{90D2B1}0.09 &
  \cellcolor[HTML]{A5DAC0}0.09 &
  \cellcolor[HTML]{57BB8A}0.92 &
  \cellcolor[HTML]{86CEAB}0.25 &
  \cellcolor[HTML]{71C69C}0.63 &
  \cellcolor[HTML]{57BB8A}0.80 \\
Coarse: MLP$_\text{M}$; Local: default &
  \cellcolor[HTML]{FFFFFF}0.20 &
  \cellcolor[HTML]{FFFFFF}0.09 &
  \cellcolor[HTML]{FFFFFF}0.10 &
  \cellcolor[HTML]{8AD0AE}0.92 &
  \cellcolor[HTML]{57BB8A}0.25 &
  \cellcolor[HTML]{57BB8A}0.64 &
  \cellcolor[HTML]{5CBD8D}0.80 \\
Coarse: MLP$_\text{XL}$; Local: none &
  \cellcolor[HTML]{E67C73}0.42 &
  \cellcolor[HTML]{E67C73}0.32 &
  \cellcolor[HTML]{E67C73}0.40 &
  \cellcolor[HTML]{E67C73}0.65 &
  \cellcolor[HTML]{E67C73}0.05 &
  \cellcolor[HTML]{E67C73}0.22 &
  \cellcolor[HTML]{E67C73}0.37
\end{tabular}\vspace{-2mm}
\caption{\label{tab:ablation_coarse_fine}Ablation study for coarse and local refinement. 
Our used configuration is highlighted is {\color{blue}blue}.
}\vspace{-2mm}
\end{table}

Tab.~\ref{tab:ablation_coarse_fine} shows that employing just a large MLP$_\text{XL}$ (four hidden layers, width 128, $m$=6, $k$=16 as defined in Section~\ref{sec:coarse_refinement}) with the suggested losses, but without a per-vertex local refinement, resulted in poor performance. 
The same applies to bypassing the coarse alignment.

An MLP$_\text{M}$ with two 32 channel wide hidden layers ($m$=3, $k$=5), MLP$_\text{S}$ with two  16 channel wide hidden layers ($m$=3, $k$=5), or just global $(o^g,s^g)$ used as optimized parameters during coarse refinement yielded similar outcomes. 
Relying solely on $(o^g,s^g)$ leads to low distance-based error but poor accuracy (e.g., $\tau=0.1$), revealing alignment precision issues. 
MLP$_\text{M}$ showed higher distance errors but improved accuracy, suggesting overfitting to sparse data with excessive adjustments in less populated areas. 
MLP$_\text{S}$ achieved a balance, effectively merging precision and robustness.

\paragraph*{Mesh Decimation.}

In Main-Sec.~\ref{sec:meshingsec}, introduced the intial downample factor $d$ for the triangulation and the quadric mesh decimation ratio $r$. 
In Main-Tab.~\ref{tab:ablation_every_step}, we show that the meshing step, where $d$=4 was used, has an negligible impact on the depth map. 
Using $d$ = 4 reduced the 1738k values in the depth map to 224k vertices (87\,\% reduction).
Additionally, we evaluated with $d$ = 8 and different mesh decimation ratios $r$, as shown in Tab.~\ref{tab:ablation_decimation}.

Setting the initial downsample factor to $d$ = 4 showed consistently better results than $d$ = 8.
Mesh decimation ratios $r$ of 1.0, 0.5 (reducing to 112k vertices), and 0.25 (56k vertices) show very negligible differences, with 0.5 slightly outperforming 1.0 on some metrics.
Based on this, we chose $r$= 0.5 as the default.

\begin{table}[]
\centering
\tiny
\begin{tabular}{l|rrrr|rrr}
Setup &
  \multicolumn{1}{l}{RMSE} &
  \multicolumn{1}{l}{MAE} &
  \multicolumn{1}{l}{L1-rel} &
  \multicolumn{1}{l|}{L1-inv} &
  \multicolumn{1}{c}{Acc.$\uparrow$} &
  \multicolumn{1}{c}{Acc.} &
  \multicolumn{1}{c}{Acc.} \\
   & \multicolumn{1}{c}{$\downarrow$}
  &
  &
  &
 &
  \multicolumn{1}{c}{0.01\,m} &
  \multicolumn{1}{c}{0.05\,m} &
  \multicolumn{1}{c}{0.10\,m} \\\hline
  $d= 4$, $r =1$ &
  \cellcolor[HTML]{A5DAC1}0.203 &
  \cellcolor[HTML]{67C195}0.088 &
  \cellcolor[HTML]{6DC499}0.091 &
  \cellcolor[HTML]{57BB8A}0.078 &
  \cellcolor[HTML]{57BB8A}0.26 &
  \cellcolor[HTML]{57BB8A}0.65 &
  \cellcolor[HTML]{57BB8A}0.82 \\
{\color{blue}$d= 4$, $r =0.5$} &
  \cellcolor[HTML]{57BB8A}0.198 &
  \cellcolor[HTML]{57BB8A}0.087 &
  \cellcolor[HTML]{57BB8A}0.090 &
  \cellcolor[HTML]{71C59C}0.080 &
  \cellcolor[HTML]{81CCA8}0.25 &
  \cellcolor[HTML]{88CFAD}0.64 &
  \cellcolor[HTML]{7ECBA5}0.81 \\
$d= 4$, $r =0.25$ &
  \cellcolor[HTML]{99D6B8}0.202 &
  \cellcolor[HTML]{96D4B6}0.090 &
  \cellcolor[HTML]{ABDDC4}0.095 &
  \cellcolor[HTML]{B2E0C9}0.084 &
  \cellcolor[HTML]{A0D9BD}0.25 &
  \cellcolor[HTML]{ABDDC5}0.63 &
  \cellcolor[HTML]{B5E1CC}0.80 \\
$d= 8$, $r =1$ &
  \cellcolor[HTML]{EFA9A3}0.213 &
  \cellcolor[HTML]{F0AFA9}0.098 &
  \cellcolor[HTML]{F2B7B2}0.106 &
  \cellcolor[HTML]{F4C4C0}0.095 &
  \cellcolor[HTML]{F0B3AE}0.23 &
  \cellcolor[HTML]{F1BAB5}0.60 &
  \cellcolor[HTML]{F3C2BE}0.77 \\
$d= 8$, $r =0.5$ &
  \cellcolor[HTML]{ED9D96}0.214 &
  \cellcolor[HTML]{EC9992}0.100 &
  \cellcolor[HTML]{E98880}0.109 &
  \cellcolor[HTML]{EC9790}0.099 &
  \cellcolor[HTML]{F0B5B0}0.23 &
  \cellcolor[HTML]{EC9B95}0.59 &
  \cellcolor[HTML]{F0B4AF}0.77 \\
$d= 8$, $r =0.25$ &
  \cellcolor[HTML]{E67C73}0.216 &
  \cellcolor[HTML]{E67C73}0.101 &
  \cellcolor[HTML]{E67C73}0.110 &
  \cellcolor[HTML]{E67C73}0.101 &
  \cellcolor[HTML]{E67C73}0.22 &
  \cellcolor[HTML]{E67C73}0.58 &
  \cellcolor[HTML]{E67C73}0.76                   
\end{tabular}
\caption{Ablation study on downsampling factor~$d$ and decimation ration~$r$. A higher value of $d$ indicates lower geometric resolution, while a higher value of $r$ indicates a higher geometric resolution.}
    \label{tab:ablation_decimation}

\end{table}

\paragraph*{Coordinate Refinement.}

\begin{figure}
    \centering
    \includegraphics[width=0.99\linewidth]{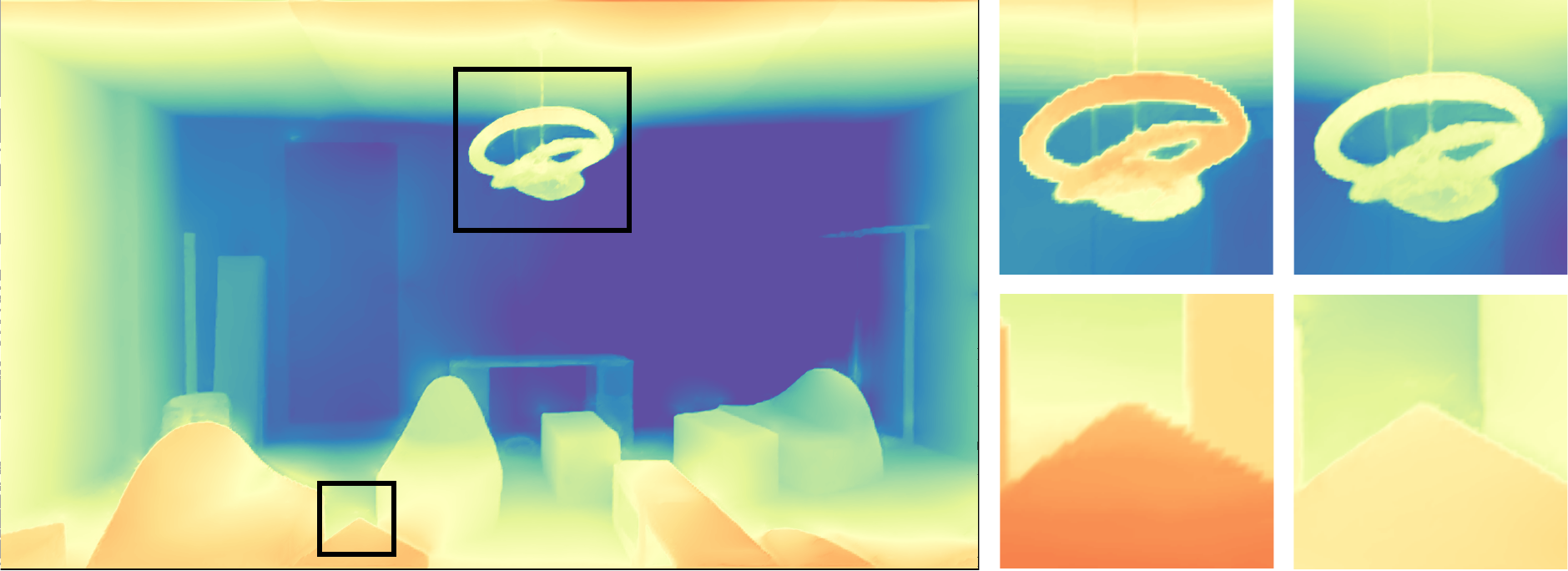}
    \caption{Refined edges through $(u,v)$ coordinate refinement. Left crops: Initial depth map exhibits jaggy edges along silhouettes. Right crops: After convergence, silhouettes are piece-wise linearly approximated by triangle edges. }
    \label{fig:uv_refinement}
\end{figure}

Fig.~\ref{fig:uv_refinement} shows the influence of the $(u,v)$ coordinate refinement, as introduced in Main-Sec.~\ref{sec:local_refinement}, thus counterbalancing the initial resolution loss by setting $d$ = 4.
This refinement significantly improved sharpness and accuracy in our depth maps, see Tab.~\ref{tab:ablation_losses}.

\paragraph*{Initial Depth Estimation.}
\begin{table}[]
\centering
\tiny
\addtolength{\tabcolsep}{-2pt}
\begin{tabular}{l|rrrr|rrrr}
Setup &
  \multicolumn{1}{c}{RMSE} &
  \multicolumn{1}{c}{MAE} &
  \multicolumn{1}{c}{L1-rel} &
  \multicolumn{1}{c|}{L1-inv} &
  \multicolumn{1}{c}{$\delta_1$} &
  \multicolumn{1}{c}{Acc.} &
  \multicolumn{1}{c}{Acc.} &
  \multicolumn{1}{c}{Acc.} \\
   & \multicolumn{1}{c}{$\downarrow$}
  &
  &
   &
  &\multicolumn{1}{c}{$\uparrow$}
 &
  \multicolumn{1}{c}{0.01\,m} &
  \multicolumn{1}{c}{0.05\,m} &
  \multicolumn{1}{c}{0.10\,m} \\
  \hline
Marigold &
  \cellcolor[HTML]{64C093}0.18 &
  \cellcolor[HTML]{57BB8A}0.09 &
  \cellcolor[HTML]{57BB8A}0.09 &
  \cellcolor[HTML]{76C7A0}0.10 &
  \cellcolor[HTML]{57BB8A}0.92 &
  \cellcolor[HTML]{57BB8A}0.25 &
  \cellcolor[HTML]{57BB8A}0.63 &
  \cellcolor[HTML]{57BB8A}0.79 
   \\
DepthAnything2$_\textrm{S}$ &
  \cellcolor[HTML]{F1B4AF}0.21 &
  \cellcolor[HTML]{F7D5D2}0.10 &
  \cellcolor[HTML]{FAE1DF}0.11 &
  \cellcolor[HTML]{E67C73}0.13 &
  \cellcolor[HTML]{FEFAF9}0.91 &
  \cellcolor[HTML]{EFF9F4}0.21 &
  \cellcolor[HTML]{E3F4EC}0.56 &
  \cellcolor[HTML]{DDF2E8}0.75 
   \\
DepthAnything2$_\textrm{L}$ &
  \cellcolor[HTML]{57BB8A}0.18 &
  \cellcolor[HTML]{BDE4D1}0.09 &
  \cellcolor[HTML]{5EBE8F}0.09 &
  \cellcolor[HTML]{57BB8A}0.10 &
  \cellcolor[HTML]{F7FCF9}0.91 &
  \cellcolor[HTML]{F6D2CF}0.20 &
  \cellcolor[HTML]{F3C5C1}0.53 &
  \cellcolor[HTML]{F5CDC9}0.72 
   \\
DepthAnything2$_\textrm{Metric}$ &
  \cellcolor[HTML]{E67C73}0.22 &
  \cellcolor[HTML]{E67C73}0.11 &
  \cellcolor[HTML]{E67C73}0.16 &
  \cellcolor[HTML]{EEA29C}0.13 &
  \cellcolor[HTML]{E67C73}0.90 &
  \cellcolor[HTML]{E67C73}0.19 &
  \cellcolor[HTML]{E67C73}0.51 &
  \cellcolor[HTML]{E67C73}0.70 
\end{tabular}
\caption{\label{tab:init_dm_abl}Monocular depth estimator usage in our pipeline.}
\end{table}

Comparing different monocular depth estimation techniques (see Tab.~\ref{tab:init_dm_abl}), both Marigold and Depth Anything2 Large (L) perform nearly equally well.
This suggests our method might benefit from future advancements in monocular estimators.
Furthermore, although the DepthAnything2 Small (S) model produces results in seconds compared to minutes for Marigold, its performance is only marginally lower than its larger counterparts. 
In case of DepthAnything2 metric, we did not apply our global scale estimation and instead relied on the absolute values of the depth map.
The qualitative differences indicate that metric prediction was still comparably inaccurate which could not be outweighed by later optimization steps..
Based on the overall slight lead, we chose Marigold as the default for our method if not indicated otherwise.

\paragraph*{Tonemapping.}
\begin{figure}
    \centering
    \includegraphics[width=0.98\linewidth]{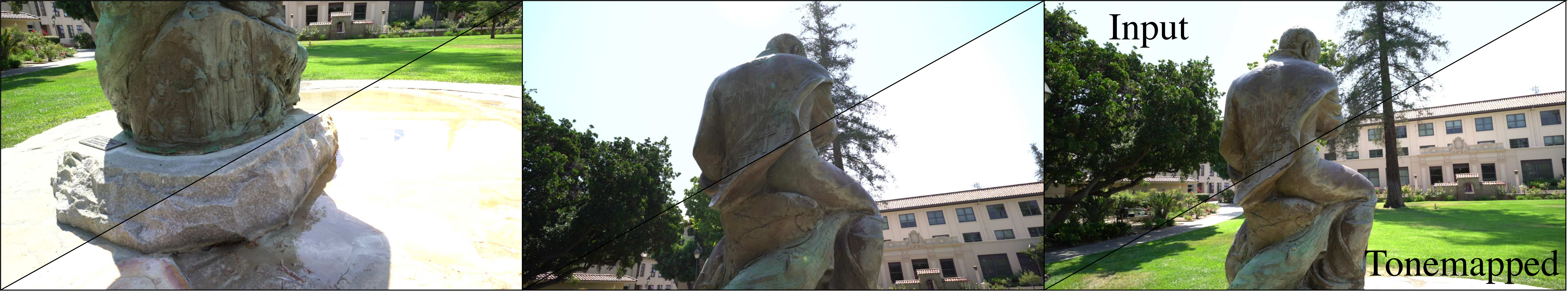}
    \caption{Tonemapping. In datasets with high variance in exposures (such as Tanks and Temples Ignatius), our tonemapping module still allows photometric optimization. }
    \label{fig:tonemapping}
\end{figure}

Tonemapping improves results for datasets like Ignatius which exhibit high variances of camera parameters like exposure or response.
In this case, precision went up from 0.52 to 0.55, recall from 0.77 to 0.86, F-score from 0.62 to  0.67.

An example of the impact on the input images is shown in Fig.~\ref{fig:tonemapping}.
Such balanced images could also be used to produce more evenly colored point clouds.

\paragraph*{Dense Geometric Consistency \& Multi Depth Map Refinement.}
Moreover, we explored extending our technique to concurrently optimizing multiple depth maps. 
To do so, we selected additional views of the image set $I_n$ to predict depth maps.
Then, we follow our global estimation and meshing procedure for additional views of the image set $I_n$ to get intial depth maps.
During their concurrent refinement, we 1) add a geometric consistency loss $\lossgeomconsistency$  that enforces that projected depth values that have similar color also should have equal depth and 2) employ a simple novel view synthesis algorithm that does alpha composition of the warped depth maps to better resolve visibility (instead of masking) to calculate another photometric loss $\lossmse^\text{MV}$.

As shown in Tab.~\ref{tab:ablation_multiview}, qualitative differences are minor and we could not determine noteworthy benefits.
However, the number of optimizable parameters scales with the number of depth maps and the introduced losses expand the computational graph for backpropagation further. 
Consequently, the required VRAM increases drastically (possible batch size had to be reduced from 14 to 4 in case of 24\,GB VRAM),
so we decided against further usage.

\begin{table}[]
\centering
\tiny
\addtolength{\tabcolsep}{-2pt}
\begin{tabular}{l|rrrr|rrrr}
Setup &
  \multicolumn{1}{c}{RMSE} &
  \multicolumn{1}{c}{MAE} &
  \multicolumn{1}{c}{L1-rel} &
  \multicolumn{1}{c|}{L1-inv} &
  \multicolumn{1}{c}{$\delta_1$} &
  \multicolumn{1}{c}{Acc.} &
  \multicolumn{1}{c}{Acc.} &
  \multicolumn{1}{c}{Acc.} \\
   & \multicolumn{1}{c}{$\downarrow$}
  &
  &
   &
  &\multicolumn{1}{c}{$\uparrow$}
 &
  \multicolumn{1}{c}{0.01\,m} &
  \multicolumn{1}{c}{0.05\,m} &
  \multicolumn{1}{c}{0.10\,m} \\
  \hline
$\scriptstyle \lossmse$ &
  \cellcolor[HTML]{E67C73}0.194 &
  \cellcolor[HTML]{FEF6F5}0.091 &
  \cellcolor[HTML]{FAE2E0}0.095 &
  \cellcolor[HTML]{EEF8F3}0.112 &
  \cellcolor[HTML]{F5CBC8}0.906 &
  \cellcolor[HTML]{57BB8A}0.262 &
  \cellcolor[HTML]{57BB8A}0.605 &
  \cellcolor[HTML]{F7D6D3}0.760 \\
$\scriptstyle \lossmse^\text{MV}$ &
  \cellcolor[HTML]{57BB8A}0.175 &
  \cellcolor[HTML]{57BB8A}0.088 &
  \cellcolor[HTML]{57BB8A}0.093 &
  \cellcolor[HTML]{FFFFFF}0.112 &
  \cellcolor[HTML]{57BB8A}0.920 &
  \cellcolor[HTML]{E98E86}0.193 &
  \cellcolor[HTML]{F4C5C1}0.584 &
  \cellcolor[HTML]{57BB8A}0.777 \\
$\scriptstyle \lossgeomconsistency$ &
  \cellcolor[HTML]{94D3B4}0.179 &
  \cellcolor[HTML]{F7FCFA}0.091 &
  \cellcolor[HTML]{E67C73}0.097 &
  \cellcolor[HTML]{57BB8A}0.111 &
  \cellcolor[HTML]{5BBD8D}0.920 &
  \cellcolor[HTML]{E67C73}0.188 &
  \cellcolor[HTML]{E67C73}0.578 &
  \cellcolor[HTML]{8ED2B1}0.773 \\
$\scriptstyle \lossmse + \lossgeomconsistency+  \lossmse^\text{MV}$ &
  \cellcolor[HTML]{EA918A}0.193 &
  \cellcolor[HTML]{E67C73}0.093 &
  \cellcolor[HTML]{C5E7D7}0.094 &
  \cellcolor[HTML]{E67C73}0.129 &
  \cellcolor[HTML]{E67C73}0.895 &
  \cellcolor[HTML]{5FBF90}0.260 &
  \cellcolor[HTML]{D2EDE0}0.593 &
  \cellcolor[HTML]{E67C73}0.745
\end{tabular}
    \caption{Ablation study for a multiview setup. 
    }
    \label{tab:ablation_multiview}
\end{table}

\end{document}